\documentclass[10pt,twocolumn,letterpaper]{article}

\usepackage[pagenumbers]{cvpr} 

\usepackage{cuted}

\usepackage[utf8]{inputenc} 
\usepackage[T1]{fontenc}    
\usepackage{url}            
\usepackage{booktabs}       
\usepackage{amsfonts}       
\usepackage{nicefrac}       
\usepackage{microtype}      
\usepackage{xcolor}         

\usepackage{hyperref}
\usepackage{graphicx}
\usepackage{caption}
\usepackage{subcaption}
\usepackage{amsthm}
\usepackage{amssymb}
\usepackage[capitalize]{cleveref}
\usepackage[breakable,skins]{tcolorbox}
\usepackage{float}
\usepackage{xcolor}
\usepackage{tikz}
\usetikzlibrary{positioning,calc, fit}

\usepackage{tabularx} 
\usepackage{xltabular}
\usepackage{multirow} 
\usepackage{wrapfig}

\usepackage{longtable}  
\usepackage{array}      

\definecolor{cvprblue}{rgb}{0.21,0.49,0.74}

\def\confName{CVPR}
\def\confYear{2027}

\title{PLAICraft: Large-Scale Time-Aligned Vision-Speech-Action Dataset for Embodied AI}

\author{
  Yingchen He \quad Christian D. Weilbach \quad Martyna E. Wojciechowska \quad Yiyuan Sun \\
    Jason Yoo \quad Yuxuan Zhang \quad Oluwanifemi A. Adekanye \\ Chia-Hong Hsu \quad Vasileios Lioutas \quad Frank Wood \\[6pt]
  University of British Columbia \\
}

\begin{document}
\maketitle

\begin{strip}
  \centering
  \includegraphics[width=0.971\textwidth]{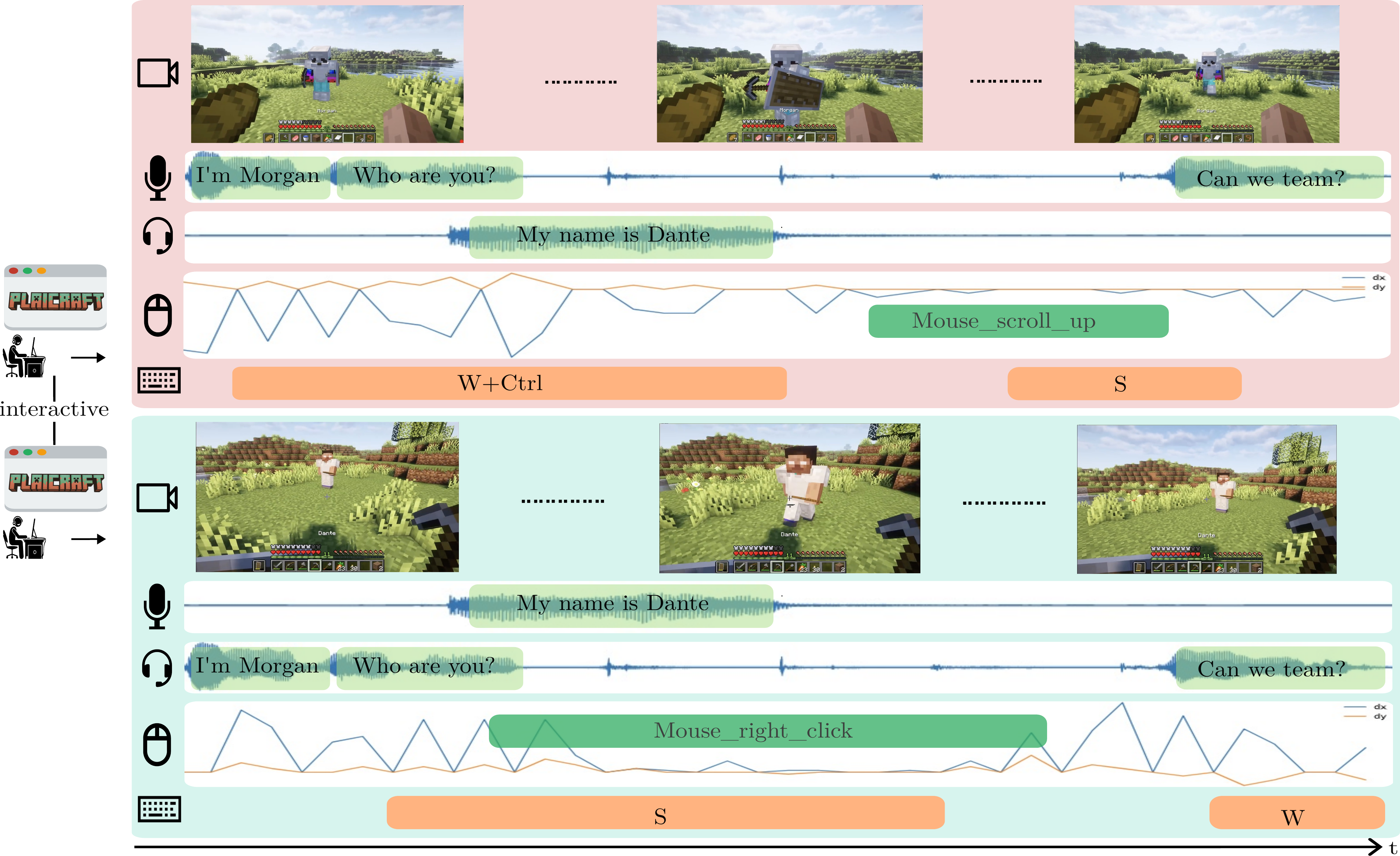}
  \captionsetup{hypcap=false}
  \captionof{figure}{
    Illustration of PLAICraft dataset collection infrastructure with two,
    but often many, Minecraft players interacting with each other through
    our instrumented gameplay environment. Players connect through a web
    browser to a virtual machine running Minecraft and data recording
    software. All interactions are real-time, and 3D proximity voice chat
    is enabled. Recordings include five time-aligned modalities: video,
    speaking audio, hearing audio, mouse and keyboard.
  }
  \captionsetup{hypcap=true}
  \label{fig:main-plot}
\end{strip}
\begin{abstract}
Advances in deep generative modeling have made it increasingly plausible to train human-level embodied agents. Yet progress has been limited by the absence of large-scale, real-time, multi-modal, and socially interactive datasets that reflect the sensory-motor complexity of natural environments. To address this, we present PLAICraft, a novel data collection platform and dataset capturing multiplayer Minecraft interactions across five time-aligned modalities: video, game output audio, microphone input audio, mouse, and keyboard actions. Each modality is logged with millisecond time precision, enabling the study of synchronous, embodied behaviour in a rich, open-ended world. The dataset comprises 5525 hours of interactive gameplay from more than 232 participants. Alongside the dataset, we provide an evaluation suite for benchmarking model capabilities in object recognition, spatial awareness, language grounding, and long-term memory. PLAICraft opens a path toward training and evaluating agents that act fluently and purposefully in real time, paving the way for truly embodied artificial intelligence.
\end{abstract}
\section{Introduction}
\label{sec:intro}

\begin{table*}[!t]
  \centering
  \tiny
  \begingroup
  \setlength{\tabcolsep}{1.25pt}
  \renewcommand{\arraystretch}{1.12}
  \everymath{\scriptstyle}
  \thinmuskip=1mu \medmuskip=2mu \thickmuskip=3mu \mathsurround=0pt
  \newcolumntype{Y}{>{\raggedright\arraybackslash}X}
  \newcolumntype{N}{>{\raggedright\arraybackslash}l}
  \begin{tabularx}{\textwidth}{@{} l | Y Y Y Y Y @{}}
    \toprule
    \textbf{} &
    \multicolumn{1}{N}{\textbf{PLAICraft}} &
    \multicolumn{1}{N}{MineRL} &
    \multicolumn{1}{N}{MineDojo} &
    \multicolumn{1}{N}{VPT Contractor} &
    \multicolumn{1}{N}{Mind2Web} \\
    \midrule
    \textbf{Domain} &
    Minecraft &
    Minecraft & Minecraft & Minecraft &
    Website GUI \\
    \midrule
    \textbf{Modalities} &
    \textbf{Video; Hearing\&Speaking audio; Mouse; Keyboard} &
    Video; Sim state; Discrete actions &
    Video; Text &
    Video; Discrete actions &
    HTML/DOM; Text \\
    \midrule
    \textbf{Scale} &
    5,525 h &
    500 h &
    $\sim$300{,}000 h &
    $\sim$2{,}000 h &
    $\sim$50 h (est. from 2{,}350 tasks) \\
    \midrule
    \textbf{Source} &
    232 global volunteers &
    Paid crowdsource &
    Internet &
    Internet \& Paid crowdsource &
    Paid crowdsource \\
    \midrule
    \textbf{\begin{tabular}[t]{@{}l@{}}Action\\Space\end{tabular}} &
    \begin{minipage}[t]{\linewidth}\raggedright\setlength{\baselineskip}{10pt}
      $m_t\!\in\!\mathbb{R}^{2}$, 100\,Hz;\\
      $k_t\!\in\!\{0,1\}^{79}$, 100\,Hz;\\
      $a^{speak}_t$:mono, 48\,kHz
    \end{minipage}
    &
    \begin{minipage}[t]{\linewidth}\raggedright\setlength{\baselineskip}{10pt}
      $u^{\mathrm{MRL}}_t\!\in\!\{0,1\}^{23}$, 20\,Hz;\\
      $c^{\mathrm{MRL}}_t\!\in\!\mathbb{R}^{2}$, 20\,Hz\\
    \end{minipage}
    &
    \begin{minipage}[t]{\linewidth}\raggedright\setlength{\baselineskip}{10pt}
      $\varnothing$
    \end{minipage}
    &
    \begin{minipage}[t]{\linewidth}\raggedright\setlength{\baselineskip}{10pt}
      $u^{\mathrm{VPT}}_t\!\in\!\{0,1\}^{20}$, 20\,Hz;\\
      $m^{\mathrm{VPT}}_t\!\in\!\mathbb{R}^{2}$, 20\,Hz
    \end{minipage}
    &
    \begin{minipage}[t]{\linewidth}\raggedright\setlength{\baselineskip}{10pt}
      $\mathrm{op}^{\mathrm{M2W}}_t\!\in\!\mathcal{O}$;\\
      $\mathrm{elem}^{\mathrm{M2W}}_t\!\in\!\mathcal{N}$;\\
      $\mathrm{val}^{\mathrm{M2W}}_t\!\in\!\mathcal{T}^{L}$
    \end{minipage}
    \\
    \midrule
    \textbf{\begin{tabular}[t]{@{}l@{}}Obser-\\vation\\Space\end{tabular}} &
    \begin{minipage}[t]{\linewidth}\raggedright\setlength{\baselineskip}{10pt}
      $v_t\!\in\!\mathbb{R}^{720\times1280\times3}$, 30\,Hz;\\
      $a^{hear}_t$:mono, 48\,kHz
    \end{minipage}
    &
    \begin{minipage}[t]{\linewidth}\raggedright\setlength{\baselineskip}{10pt}
      $v^{\mathrm{MRL}}_t\!\in\!\mathbb{R}^{64\times64\times3}$ $\lor$ $\mathbb{R}^{192\times256\times3}$, 20\,Hz;\\
      $\mathrm{game}^{\mathrm{MRL}}_t\!\in\!\mathcal{D}$,
    \end{minipage}
    &
    \begin{minipage}[t]{\linewidth}\raggedright\setlength{\baselineskip}{10pt}
      $v^{\mathrm{MDJ}}_t\!\in\!\mathbb{R}^{H\times W\times3}$;\\
      $T^{\mathrm{MDJ}}_t\!\in\!\mathcal{T}^{L}$
    \end{minipage}
    &
    \begin{minipage}[t]{\linewidth}\raggedright\setlength{\baselineskip}{10pt}
      $v^{\mathrm{VPT}}_t\!\in\!\mathbb{R}^{128\times128\times3}$, 20\,Hz
    \end{minipage}
    &
    \begin{minipage}[t]{\linewidth}\raggedright\setlength{\baselineskip}{10pt}
      $D^{\mathrm{M2W}}_t \!\in\! \mathcal{G}$;\\
      $\mathrm{instr}^{\mathrm{M2W}} \!\in\! \mathcal{T}^{L}$
    \end{minipage}
    \\
    \midrule
    \textbf{Open Task} &
    \textbf{Yes} &
    No & No & No & No \\
    \midrule
    \textbf{Reset} &
    \textbf{No} &
    Episodic & Episodic & Episodic & Episodic \\
    \midrule
    \textbf{Social} &
    \textbf{Yes} &
    No & No & No & No \\
    \bottomrule
  \end{tabularx}
  \caption{\small Comparison between PLAICraft and related datasets:
  \textit{PLAICraft sets a new benchmark for embodied-AI datasets by providing high-resolution video data, high-frequency continuous keyboard \& mouse data and speaking \& hearing audio data at an immense scale.}
  \textbf{PLAICraft}: $v_t$:pov video; $a^{speak}_t$:speaking audio; $a^{hearing}_t$:hearing audio; $m_t$:mouse movement; $k_t$:keyboard presses+mouse clicks/scrolls.
  \textbf{MineRL}: $v^{\mathrm{MRL}}_t$:pov video; $\mathrm{game}^{\mathrm{MRL}}_t$ a set of game-state(player inventory, events, distance to objects...) $u^{\mathrm{MRL}}_t$:discrete action(attack, forward, drop,...); $c^{\mathrm{MRL}}_t$:camera pitch\&yaw.
  \textbf{MineDojo}: $v^{\mathrm{MDJ}}_t$:YouTube video, resolution and frame rate vary; $T^{\mathrm{MDJ}}_t$:transcript tokens; $\mathcal{T}^{L}$:a sequence of text tokens.
  \textbf{VPT}: $v^{\mathrm{VPT}}_t$:video; $u^{\mathrm{VPT}}_t$:discrete actions; $c^{\mathrm{VPT}}_t$:coarse mouse movement.
  \textbf{Mind2Web}: $D^{\mathrm{M2W}}_t$:HTML/DOM graph; $\mathrm{instr}^{\mathrm{M2W}}$:text instruction; $\mathrm{op}^{\mathrm{M2W}}_t$:operation, $\mathcal{O}=\{CLICK, TYPE, SELECT\}$; $\mathrm{elem}^{\mathrm{M2W}}_t$:target element, $\mathcal{N}:$DOM elements; $\mathrm{val}^{\mathrm{M2W}}_t$:text value.}
  \label{tab:plaicraft-comparison}
  \endgroup
\end{table*}

Embodied artificial intelligence (EAI) aims to build agents that perceive, act, and learn in dynamic environments through real-time sensory-motor interaction. As \citet{brooks1991intelligence} emphasized, such agents must ``cope appropriately and in a timely fashion with changes in [their] dynamic environment'' and ``do something in [their] world; [they] should have some purpose in being.'' These principles remain vital as research shifts toward agentic AI systems like Devin \citep{devin2024} and Claude Code \citep{anthropic2024claude} that operate autonomously across extended tasks. 

Yet most modern agents are \textit{disembodied} in the sense that they interact with their environment indirectly, often through high-level text interfaces and asynchronous API calls. This limits their ability to engage in temporally grounded, perceptual-motor loops that characterize embodied intelligence. While simulated environments such as Habitat \citep{savva2019habitat}, RoboTHOR \citep{deitke2020robothor}, and ThreeDWorld \citep{gan2021threedworld} support real-time, sensorimotor learning, they often lack socially interactive or linguistically rich scenarios. Minecraft datasets like MineRL \citep{guss2019minerl}, OpenAI’s VPT \citep{baker2022videopretrainingvptlearning}, and CraftAssist \citep{das2019tacl} have driven progress in perception–action coupling and language learning. More recently, MineDojo \citep{fan2022minedojo} introduced large-scale web-mined video-language data to support open-ended instruction following and knowledge transfer. However, these settings remain offline, non-interactive, heterogeneous, or lack real-time, social \emph{grounding}—that is, the process of linking symbols to perceptual input, physical action, or shared environmental context.

We introduce \textbf{PLAICraft}, a dataset and platform designed for the study of real-time, socially situated, speech-interactive embodied agents. Built atop a modded multiplayer Minecraft server with a proximity-based voice chat plugin, PLAICraft records time-aligned streams of screen video, game output audio, player microphone input audio, keyboard, and mouse input with millisecond precision. Agents operate in persistent, open-ended worlds with other players where perception, speech, and action are tightly coupled in time.  In its emphasis on rich, temporally grounded sensory-motor interaction, PLAICraft aligns with recent efforts to build datasets that support embodied learning in complex, naturalistic domains—such as autonomous driving benchmarks designed to capture diverse, real-world scenarios \citep{waymo2025e2e, zurn2024wayvescenes101}. 

\cref{fig:main-plot} illustrates our approach: participants play in our virtual environment and interact with other players naturally while all modalities are recorded. Unlike most prior datasets, PLAICraft emphasizes dialogue, social context, and temporal embodiment, supporting the study of skills like object permanence, reactive speech, memory, and social reasoning. Agents must act with purpose amid uncertainty, social influence, and ambiguity \citep{Voudouris2022-VOUDHC}. PLAICraft, like the real world, lacks explicit, global reward signals and is not structured around predefined episodic tasks. While the game mechanics do allow for death and respawn cycles, the agent persists through its avatar’s deaths, and such events may even form part of emergent strategies or social play.


Our dataset extends embodied AI research beyond social speech interactive agents. For example, the application-agnostic architecture of our platform facilitates extending Minecraft agents into general computer-use agents, as our action spaces are raw keyboard, mouse and speaking audio that can be used in any applications. Furthermore, the persistent world and agent states are well-suited to continual learning \citep{Kirkpatrick_2017, li2017learningforgetting, wang2024comprehensivesurveycontinuallearning, yoo2024lifelonglearningvideodiffusion}, which requires long-horizon memory over extended gameplay. Multi-perspective participant recordings enable the systematic evaluation of third-person learning \citep{stadie2019thirdpersonimitationlearning}, training agents strictly from offline behavioural logs rather than on-policy experience.

We have collected over 13,300 hours of data from over 18,000 players around the world, from which we have curated a high-quality dataset of \textbf{5,525} hours from \textbf{232} players for release. We introduce the PLAICraft dataset in \cref{sec:plaicraft_dataset}, the data collection platform in \cref{sec:data_collection}. \cref{sec: evaluation} covers an evaluation suite designed to probe reasoning, memory, and communicative competence. 
\section{Related Work}

\textbf{Minecraft for Machine Learning} Minecraft serves as an important tool for machine learning research \citep{lifshitz2024steve1,Zhou2024MineDreamerLT}. Numerous datasets and benchmarks have been established on Minecraft \citep{guss2019minerl,milani2023bedd,fan2022minedojo,baker2022videopretrainingvptlearning}. Microsoft’s \emph{MALMO} platform \citep{10.5555/3061053.3061259} first provided a programmable interface for RL in Minecraft; building on this, \emph{MineRL} \citep{guss2019minerl,shah2021minerl} introduced human-demonstration datasets and benchmarks for control tasks. \emph{MineDojo} \citep{fan2022minedojo} extended this line by aggregating web-scale gameplay videos with language, while the \emph{VPT Contractor} dataset \citep{baker2022videopretrainingvptlearning} contributed single-player demonstrations with aligned video and keyboard/mouse actions used to train inverse-dynamics and behavioural-cloning foundation models. The \emph{BASALT} competition \citep{shah2021minerl} applied human evaluation to open-ended tasks, and \emph{BEDD} \citep{milani2023bedd} further systematized evaluation across diverse, complex Minecraft scenarios.\\
\textbf{EAI Datasets in Other Domains}
Beyond Minecraft, several large-scale datasets support embodied and interactive agents in other settings. \emph{GameNGen} trains on a 900M-frame corpus of VizDoom agent self-play trajectories, providing dense video–action sequences that let a diffusion model learn a real-time game engine over long horizons \citep{valevski2025diffusionmodelsrealtimegame}. \emph{Mind2Web} contributes crowdsourced interaction traces over 2{,}350 open-ended tasks on 137 real websites, using HTML/DOM state and natural-language instructions to train and evaluate generalist web agents on real UIs rather than simulators \citep{deng2023mind2web}. \emph{Waymo Open} provides synchronized multi-camera images, LiDAR, and HD maps for thousands of urban and suburban driving segments, and serves as a standard benchmark for perception and motion forecasting in autonomous driving \citep{sun2020waymo}.


We specify and compare a few of these datasets side by side with PLAICraft in \Cref{tab:plaicraft-comparison}.\\
\textbf{EAI Evaluation}
Evaluating EAI agent performance in complex environments effectively requires direct comparisons between human and AI abilities. There is a growing literature on such benchmarks, including the \emph{Animal-AI Environment} framework which has been used to compare human and AI performance on cognitive tasks inspired by animal psychology, providing insights into how agents reason and generalize \citep{Voudouris2022-VOUDHC}. Similarly, benchmarks such as the \emph{Baby Intuitions Benchmark (BIB)} assess agents' ability to infer goals, preferences, and the actions of others, highlighting gaps in AI's understanding of intuitive physics and social reasoning compared to humans \citep{gandhi2022babyintuitionsbenchmarkbib}. These approaches underscore the challenge of developing evaluation frameworks that align with human cognitive benchmarks \citep{lake2016buildingmachineslearnthink}.

\section{PLAICraft Dataset}\label{sec:plaicraft_dataset}
\begin{figure*}[!t]
  \centering
  \includegraphics[width=0.95\textwidth]{assets/figs/dataset_dynamics_plot.pdf}
  \caption{
    Visualization of the complex dynamics in the PLAICraft dataset. The frames here are a visualization that combines all modalities, not the actual data format, which stores each modality separately. Keyboard clicks are overlayed in the top left corner, and mouse clicks are overlayed in the top right corner. In the middle of the frame, the mouse movements are visualized by light blue arrows. To visualize speaking and hearing audio, we show their corresponding transcript at the bottom of each frame. 
  }
  \label{fig:dataset-dynamic-plot}
\end{figure*}
\begin{figure}[t]
  \centering
  \includegraphics[width=\linewidth]{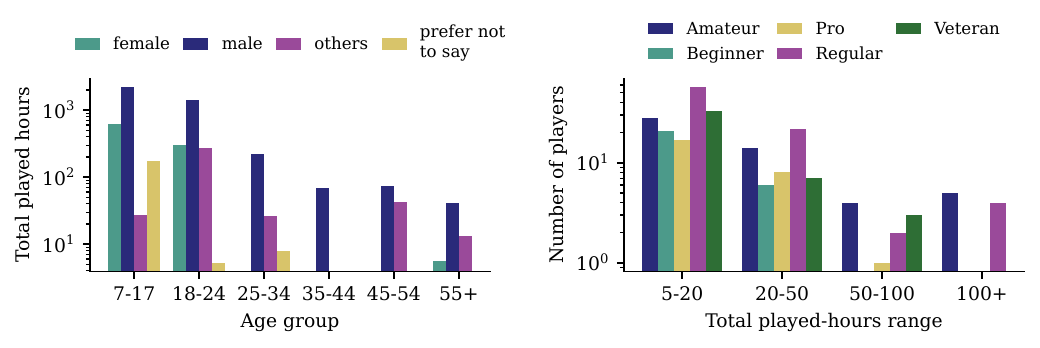}
  \caption{Self-reported participants' demographic distributions. \textbf{Left:} hours played by gender and age group. \textbf{Right:} player count by played-hours range and experience level.}
  \label{fig:dataset-statistic}
\end{figure}

We have chosen Minecraft as our environment, like many previous datasets \citep{shah2021minerl, fan2022minedojo, das2019tacl}, because its rich, open-ended dynamics make learned behaviours transferable to real-world scenarios. In vanilla Minecraft, players navigate procedurally generated 3D worlds abundant with diverse terrains, resources to gather, tools to create, structures to construct, and novel discoveries to encounter. The world players interact with is also boundary-free, allowing unrestricted exploration.  Minecraft provides neither explicit goals nor predefined storylines, rendering it uniquely suitable as a foundation for open-ended embodied AI research. Unlike many prior Minecraft datasets, which are mostly restricted to single-player mode with limited sets of tasks \citep{shah2021minerl}, limited modalities or non-systematic internet-fetched videos \citep{fan2022minedojo}, our PLAICraft dataset is unique in the following ways.

\textbf{Unbounded Multiplayer Social Environment:} The open-task nature of Minecraft inherently offers an unlimited spectrum of possible tasks. Our dataset further enhances this by employing a multiplayer environment where participants are encouraged to socially interact with each other, and a proximity-based voice chat plugin for natural voice communication. This setup substantially enriches the complexity of behaviours, particularly enabling realistic social interactions. Figure \ref{fig:dataset-dynamic-plot} shows a wide range of solo and collaborative activities players have performed on the server.

\begin{figure}[!htbp]
  \centering

  \begin{subfigure}[t]{\linewidth}
    \centering
    \includegraphics[
      width=\linewidth,
      keepaspectratio
    ]{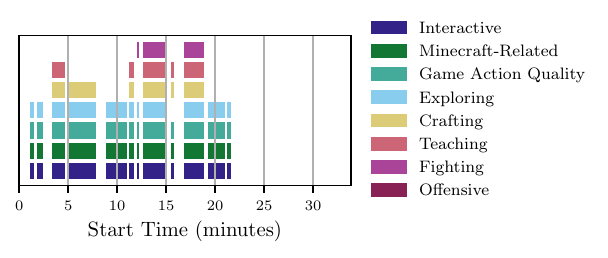}
    \caption{}
    \label{fig:interactive_segments_labels}
  \end{subfigure}

  \par\vspace{2mm}

  \begin{subfigure}[t]{\linewidth}
    \centering
    \includegraphics[
      width=\linewidth,
      keepaspectratio
    ]{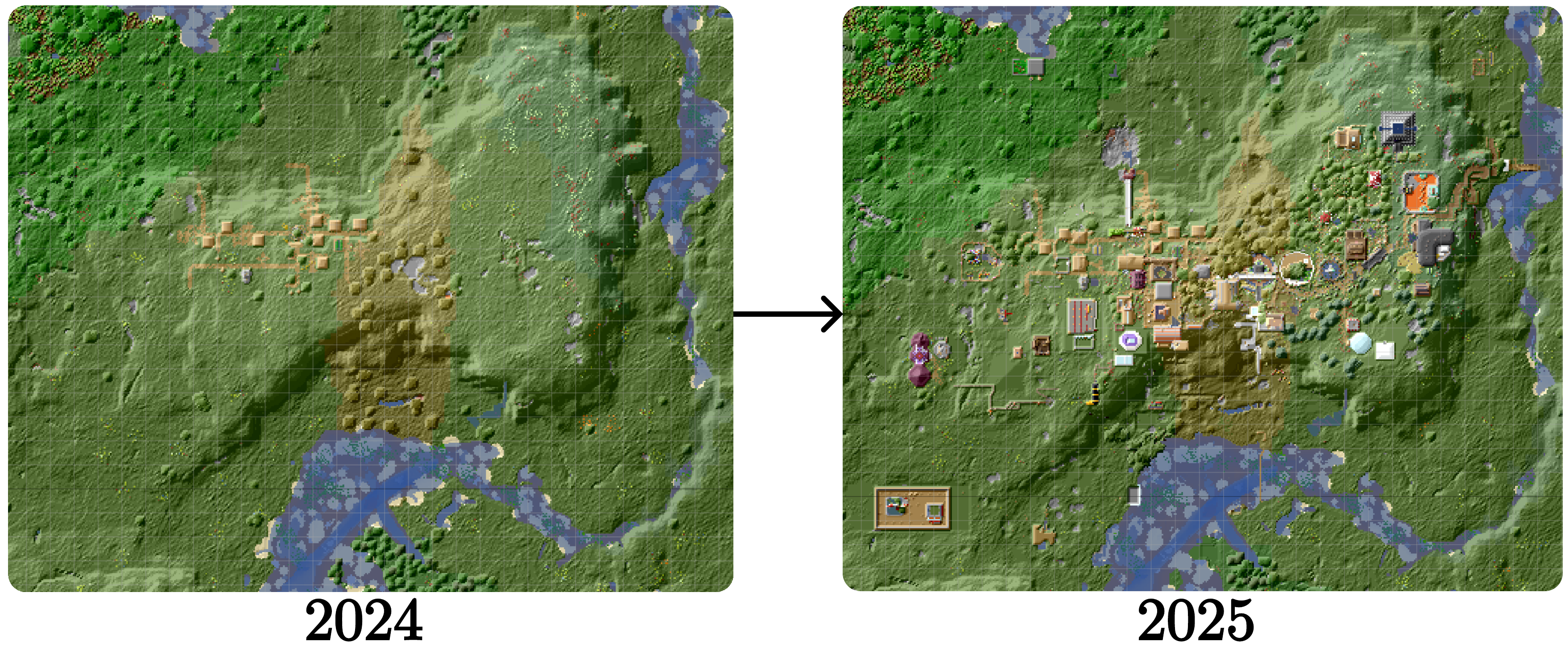}
    \caption{}
    \label{fig:player-base-evolution-plot}
  \end{subfigure}

  \caption{
    \textbf{(a)} Labelled interactive segments within a session
    using an LLM on audio transcripts
    (\cref{sec:automatic_data_annotation}).
    \textbf{(b)} Bird's-eye-view images showing world changes.
  }
  \label{fig:dataset_examples}
\end{figure}

During each gameplay session, participants receive no specific task assignments or restrictions regarding interaction areas. Instead, they are encouraged to play naturally, thinking of this not as a research project but just regular Minecraft, subject only to basic behavioural guidelines promoting mutual respect. The virtual microphone on the VM is forced to be enabled, and we encourage our players to continuously talk, especially when they play interactively with others. In each gameplay session, players have a maximum of 4 hours of play time, and they can resume their game state in the next session; thus, they can perform a wide range of different activities. For example, as shown in \Cref{fig:interactive_segments_labels}, the player engaged in a wide range of interactive activities from minute 1 to minute 22, and from minute 22 to the end, the player only performed solo activities.

Consequently, our dataset exhibits extensive dynamic variability. Solo player activities span a wide spectrum, beginning with basic actions such as moving with \texttt{WASD} keys or breaking a dirt block and progressing to moderately complex behaviours like mining, crafting, and combating. They extend further to events that require advanced skills, including defeating the Ender Dragon, constructing intricate redstone mechanisms, or even exploiting server bugs for personal benefit. Additionally, the dataset captures numerous social interactions, ranging from simple greetings and self-introductions to collaborative activities involving team formation and cooperative gameplay. Notably, participants often form teams spontaneously, engaging in continuous interaction through collective base construction and various other activities. Players between different teams interact with each other more cautiously and less frequently. Despite rigorous moderation efforts, occasional negative interactions such as conflicts, theft, and intentional property damage are present. Such behaviours are explicitly identified, and strong measures are implemented to prevent malicious participation.

\textbf{Millisecond-precision Time-aligned Multimodal Data:} The unfiltered PLAICraft dataset comprises tens of thousands of gameplay session recordings, each representing one session of an individual player. For every session, we simultaneously capture the screen recordings, game output audio, player microphone input audio, and mouse and keyboard actions. These modalities are recorded with millisecond-precision timestamps, facilitating precise temporal alignment across data streams. The alignment between mouse and keyboard inputs and their corresponding visual and auditory outputs is strictly causal, ensuring input actions always precede or coincide with their resultant effects. Using server-side keylogging plugins, we reduce the latency between recorded actions and their corresponding state transitions to the server-side processing delay.



\textbf{Continual global world state:} To simulate the real world, throughout the entire duration of our project, the server has maintained a single, persistent world state that has evolved continuously since its creation two years ago, without resets. Consequently, player attributes—such as position, inventories, and experience level—carry over seamlessly between their gameplay sessions, unless they die in the game, then they will lose their inventories and experience points if they are not picked up soon enough. Structures built by players endure indefinitely, and any environmental changes (for example, mined blocks) become permanent. All participants share the same global world, thus the action of one participant can be directly observed by nearby participants. Figure \ref{fig:player-base-evolution-plot} illustrates the development of a communal player base constructed collaboratively by roughly ten participants over a one-year period. Thus, our dataset can also be considered as the continual history of the world's evolution from the perspectives of every player.

\section{Data Collection}
\label{sec:data_collection}

\begin{figure*}[tbp]
  \centering
  \includegraphics[width=0.9\textwidth]{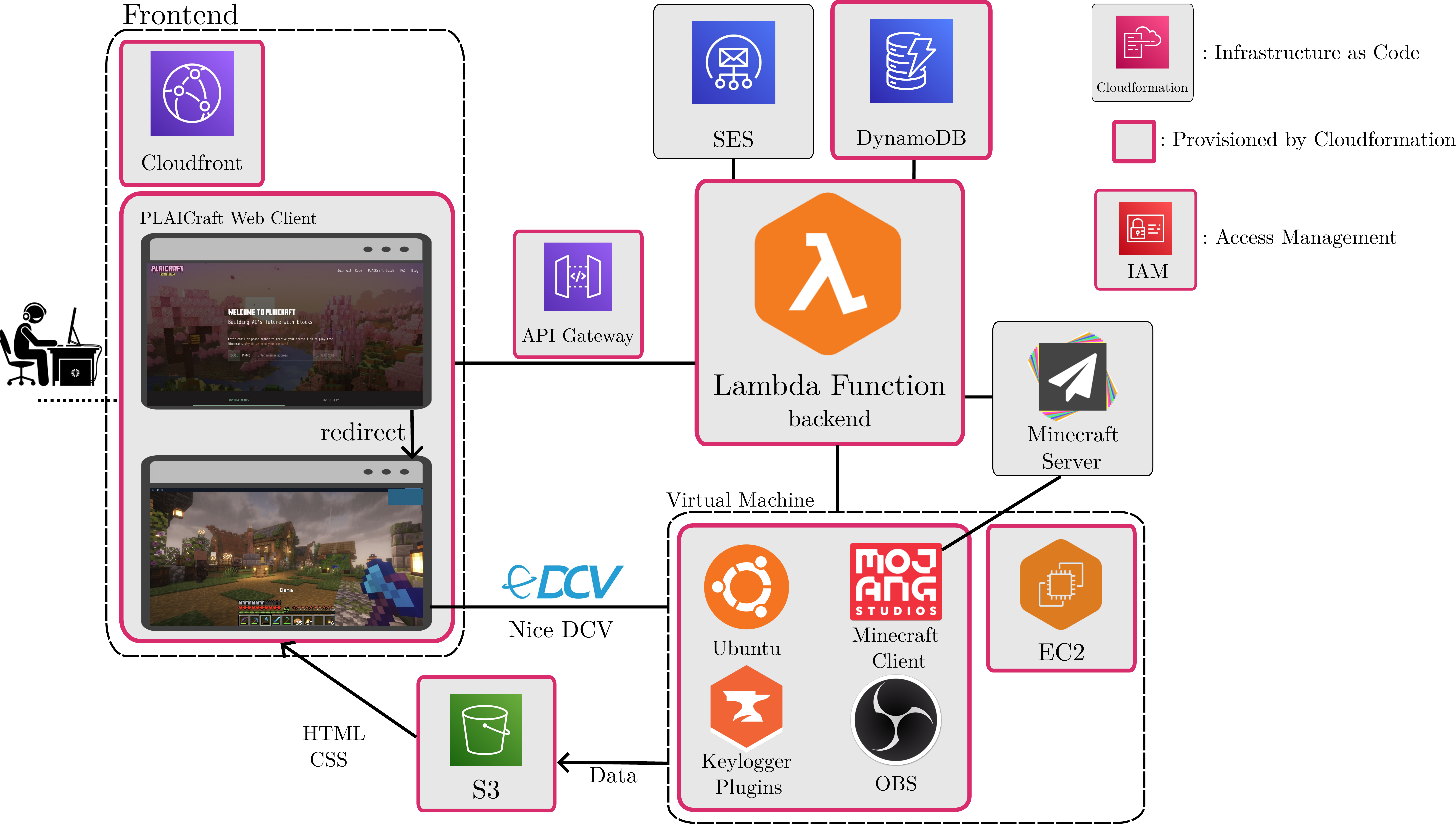}
  \caption{
    Illustration of PLAICraft data collection infrastructure. Participants can simply start a gameplay session like playing regular Minecraft, in any browser on any hardware device. Nice DCV will then connect users to a virtual machine that is running multiplayer Minecraft. On the EC2 instance, multiple pieces of recording software run in the background to record players' video, audio and mouse \& keyboard actions. Then the data will be saved to an S3 bucket. Our backend is centered around Lambda functions and multiple AWS resources.
  }
  \label{fig:aws-arch-plot}
  \vspace{-1em}
\end{figure*}

Gathering gameplay data typically requires game-specific platforms with tailored strategies for participant engagement and data collection. MineRL~\citep{guss2019minerl} uses a server and plugin system for gameplay, while MineDojo~\citep{fan2022minedojo} aggregates data from existing platforms such as YouTube, Wikipedia, and Reddit. In contrast, PLAICraft standardizes the collection of time-aligned video, game audio, microphone audio, and keyboard and mouse actions while allowing participants to engage naturally and collaboratively in multiplayer Minecraft.

\subsection{Platform Setup}
As illustrated in Figure~\ref{fig:aws-arch-plot}, our platform is built primarily on Amazon Web Services (AWS), with Python-based AWS Lambda functions serving as the backend. These functions manage AWS resources, including Amazon DynamoDB, which stores user information and gameplay status. Participants access the platform through a Chromium-compatible website hosted on Amazon S3 and delivered via CloudFront. After logging in with an email address or phone number, users are redirected to an Amazon EC2 instance for gameplay.

Each EC2 instance runs Minecraft and the associated data-collection software. We create a preconfigured Amazon Machine Image (AMI) containing all required software and settings. Upon login, each participant receives a dedicated g5.2xlarge instance launched from this AMI, ensuring secure, isolated, and consistent data recording. To maintain immersion, Minecraft runs in full-screen mode, with other UI interactions disabled. Monitoring scripts track connectivity and window focus, automatically terminating the instance and uploading recordings to S3 if the player disconnects or leaves the game window. NICE DCV streams the display to participants at up to 60 FPS and 2K resolution, providing an experience comparable to cloud gaming services such as GeForce NOW and Xbox Cloud Gaming. Since all recording occurs within the isolated virtual machine, participants' local environments remain inaccessible to the recording software, helping protect their privacy.

\subsection{Minecraft Configurations}
We preserve most vanilla Java Minecraft features, with players restricted to survival mode at normal difficulty and full access to the Overworld, Nether, and End. Teleportation and global chat facilitate social interaction, while the Simple Voice Chat plugin enables proximity-based 3D voice communication, enhancing the realism of player interactions. Our custom Autojoin plugin stores and restores player states (e.g., inventory, position, and level) through DynamoDB, ensuring continuity while allowing a limited license pool to serve non-concurrent players. This enables scaling beyond 10,000 participants. The complete plugin list is provided in \Cref{sec: Minecraft configuration}.


\subsection{Increasing Social Interactions}\label{sec3.3:social_interactions}
Social interaction is a central focus of our dataset. We foster engagement through a persistent multiplayer environment, integrated voice chat, curated mods, and web and in-game UI/UX features that facilitate collaboration and coordination. We also use WhisperX~\citep{bain2023whisperxtimeaccuratespeechtranscription} to transcribe players' speech and nearby voice chat in real time, then employ an LLM (DeepSeek-V3.2-fast~\citep{deepseekai2025deepseekv32pushingfrontieropen}) to assess the quality and quantity of social interactions. Based on these assessments, we reward players with additional solo play time to incentivize sustained social engagement.

\subsection{Data Alignment \& Pre-processing}
\begin{table}[tbp]
  \centering
  \tiny
  \setlength{\tabcolsep}{3pt}
  \renewcommand{\arraystretch}{1.1}
  \newcolumntype{Y}{>{\raggedright\arraybackslash}X}
  \begin{tabularx}{0.7\columnwidth}{@{} l l Y @{}}
    \toprule
    \textbf{Modality} &
    \textbf{Recording Software} &
    \textbf{Format} \\
    \midrule
    Video &
    OBS &
    30\,FPS, 1280$\times$720, H.264 \\
    \addlinespace[2pt]
    Speaking audio &
    OBS &
    48\,kHz, mono, AAC \\
    \addlinespace[2pt]
    Hearing audio &
    OBS &
    48\,kHz, mono, AAC \\
    \addlinespace[2pt]
    Mouse movement &
    Forge mod &
    100\,Hz, (D=2) \\
    \addlinespace[2pt]
    Key presses &
    Forge mod &
    100\,Hz, (D=79) \\
    \bottomrule
  \end{tabularx}
  \caption{Recording software and format for each modality.}
  \label{tab:modality-encoding}
\end{table}
We synchronously capture all modalities on the EC2 instance using dedicated software components, as shown in \Cref{tab:modality-encoding}. We use the exact timestamp at which OBS’s ffmpeg muxer begins writing the video file as the global start time and align all modalities to this reference. Mouse and keyboard data are categorized into mouse \& keyboard clicks and mouse movements. Click data includes discrete "PRESS" and "RELEASE" events, whereas mouse movement data captures continuous relative positional changes. 

We apply various denoising and filtering techniques to remove noisy data and harmful data. We enhance the player's microphone with RE-USE \citep{fu2026rethinkingtrainingtargetsarchitectures} to remove noise and reverberation. We then diarize it with streaming Sortformer \citep{park2025sortformernovelapproachpermutationresolved} and embed each diarized speaker with TitaNet \citep{koluguri2021titanetneuralmodelspeaker}. The player is the dominant voice across all of their sessions, and each speaker is matched to that voice at a cosine similarity of at least 0.7. We keep only the chunks in which the player is the sole active speaker and silence everything else. The kept chunks are grouped into utterances, and each utterance is transcribed with Parakeet-TDT-0.6B-v3 \citep{sekoyan2025canary1bv2parakeettdt06bv3efficient}. An utterance is discarded if its transcript contains non-ASCII characters or appears on a manually reviewed list of non-English utterances. For utterances of at least 3 s, it is also discarded if its DNSMOS P.835 \citep{reddy2022dnsmosp835nonintrusiveperceptual} overall score is 2.0 or lower. For the mouse data, Minecraft re-centres the cursor whenever an interactive GUI (crafting tables, inventory, etc.) opens or closes, and we remove those jumps from the mouse data. A reading that lands on the session's window-centre pixel is replaced by its offset from that pixel. 

\begin{table*}[t]
  \centering
  \begingroup
  \tiny
  \setlength{\tabcolsep}{2pt}
  \renewcommand{\arraystretch}{1.02}
  \everymath{\scriptstyle}
  \thinmuskip=1mu \medmuskip=2mu \thickmuskip=3mu \mathsurround=0pt
  \newcolumntype{Y}{>{\raggedright\arraybackslash}X}
  \newcolumntype{C}[1]{>{\centering\arraybackslash}p{#1}}

  \begin{tabularx}{0.9\textwidth}{@{}
    >{\raggedright\arraybackslash}p{2.45cm}
    Y
    >{\raggedright\arraybackslash}p{1.5cm}
    C{2.0cm}
    C{1.0cm}
  @{}}
    \toprule
    \textbf{Test type} &
    \textbf{Hearing Audio Within Short-term Context} &
    \textbf{Measurable} &
    \textbf{Resp. window (s)} &
    \textbf{Metric} \\
    \midrule

    audio--action binding &
    Press A. &
    A pressed. &
    689.70--689.85 &
    $\mathrm{D}_{\mathit{K}}$ \\

    spelling ability &
    Please press the letter that ends the word STOP, as in S-T-O. What letter was that? &
    P pressed. &
    115.95--116.07 &
    $\mathrm{D}_{\mathit{K}}$ \\

    object recognition &
    What do you have in your right hand? &
    Carrot. &
    510.550--511.016 &
    $\mathrm{Acc}_{\mathit{R}}$ \\

    world-model understanding &
    I’m going to shoot you with an arrow, okay? And I want you to tell me how many hearts of damage are done, okay? &
    Two. &
    482.750--483.500 &
    $\mathrm{Acc}_{\mathit{R}}$ \\

    arithmetic &
    If you add up the blocks in the wool and granite pile, how many are there? &
    Five. &
    1182.683--1183.299 &
    $\mathrm{Acc}_{\mathit{R}}$ \\

    spatial understanding &
    How many blocks away is the box? &
    Hundred. &
    583.549--584.283 &
    $\mathrm{Acc}_{\mathit{R}}$ \\
    \bottomrule
  \end{tabularx}

  \caption{\small Evaluation schema, ``prompt'' examples.   Resp. window is the timestep when the measurable, automatically assessible ``response'' occurs in seconds with respect to the start of each recording session.  Different response types require different assessment metrics.  Key presses use $\mathrm{D}_{\mathit{K}}$ Hamming distance between expected and actual binarized, binned keyboard events; spoken audio response uses $\mathrm{Acc}_{\mathit{R}}$, which should be the semantic accuracy of the response.}
  \label{tab:selected_prompt_response_pairs}
  \endgroup
\end{table*}
\subsection{Collection History \& Dataset Release}
All participants are volunteer Minecraft enthusiasts recruited globally, who typically bring greater domain experience and engage in more naturalistic interactions than paid crowdsourced workers. Over the past two years, we have collected more than 13,300 hours of data from over 18,117 participants worldwide, and the dataset continues to grow. However, the play time is unevenly distributed across participants. The median participant recorded one session of about four minutes, and 85\% of participants played for less than 30 minutes in total. Many of these participants played only briefly and did not interact with other players. To ensure the quality of the dataset, we release only the recordings of players with at least 5 hours of total play which contains 5,525 hours of gameplay recordings from 9,259 sessions. The self-reported player demographics are demonstrated in \Cref{fig:dataset-statistic}

\section{Evaluation}\label{sec: evaluation}
\subsection{Design}
\begin{figure}[!t]
  \centering
  \includegraphics[width=\linewidth]{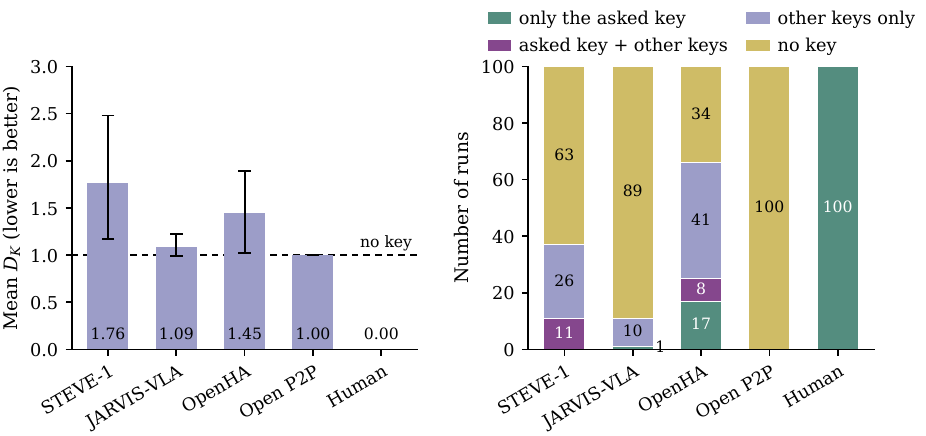}
  \caption{Open-sourced Minecraft agents on the 10 PLAICraft evaluation
keypress prompts. Each model outputs one action at the same moment the human
answered; 10 runs per prompt. Human: the player's gt recorded response. (a) Mean $D_K$, the number of missing or extra keys (lower is better), with 95\% bootstrap
confidence intervals; pressing no key scores 1. (b) Model's output keypress type of each run, very few pressed only the asked key.}
\label{fig:experiment_summary}
\end{figure}




Motivated by Animal-AI~\citep{Voudouris2022-VOUDHC}, which adopts testing paradigms from comparative psychology; BIB~\citep{gandhi2022babyintuitionsbenchmarkbib}, which assesses machines' abilities to reason about other agents' intentions by observing their actions; and the taxonomy of cognitive abilities outlined in Cattell-Horn-Carroll (CHC) theory~\citep{carroll1993human}, we publish an evaluation suite that probes how agents perceive, understand, reason, and act within PLAICraft. The idea behind this design is intuitive: if an agent truly understands what it sees and hears, it should be able to (i) talk about it, (ii) act on it, and (iii) do so within a time‑bounded, social context.

Our evaluation suite contains carefully curated ``prompts,'' where in PLAICraft a prompt is an agent evaluee's participation in a full-modality, continuous, intentional two-party session with agreement to honestly and faithfully participate elicited along with a manually labeled time-boundary that indicates when a ``prompt'' ends and the response starts.  Each such prompt is designed and labeled so as to belong to one of the ten broad cognitive abilities in CHC theory~\citep{schneider2018cattell}: fluid intelligence, quantitative knowledge, crystallized intelligence, short-term memory, visual processing, auditory processing, long-term retrieval, processing speed, decision speed, and reading/writing. Prompts, critically, are designed such that they elicit a  keyboard/mouse or verbal action that is short enough so as to not require rolling-out with a world model to evaluate.  This, by design, enables automated open-loop cognition-level policy evaluation via automatically computable metrics. 


Table~\ref{tab:selected_prompt_response_pairs} illustrates the semantic contents of six representative prompts. These prompts probe a wide range of capabilities—including visual object recognition, world-model understanding, spatial estimation, arithmetic understanding and many others. Other prompts are designed to assess cognitive abilities without requiring the  emergence of speech understanding or production.  
The full list of evaluation prompts can be found in \Cref{appendix_details_of_evaluation}.

\subsection{Experiment}
\begin{figure}[!t]
  \centering
  \includegraphics[width=\linewidth]{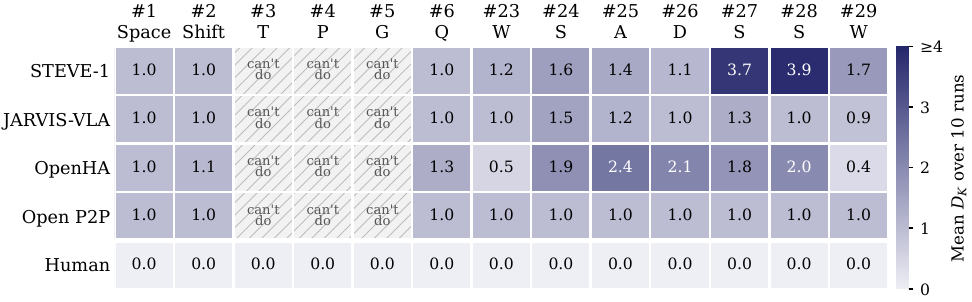}
  \caption{Mean $D_K$ over 10 runs for every key prompt. "can't do": the asked key is not in the
model's action space.}
  \label{fig:experiment_dk_per_prompt}
\end{figure}
To our knowledge, no publicly released Minecraft agent supports the full benchmark setting, which requires simultaneous visual and audio perception and responses through keyboard/mouse actions or speech. Most existing agents accept video and text instructions but lack audio input and speech output. We therefore substitute spoken instructions with transcripts, enabling evaluation on 10 of the 13 key-press prompts. The remaining three require keys (T, P, G) absent from all models' action spaces, while the other 35 prompts require verbal responses that none of the models can produce.

We evaluate four open-source agents: STEVE-1~\citep{lifshitz2024steve1}, a text-conditioned VPT policy; JARVIS-VLA~\citep{li2025jarvisvlaposttraininglargescalevision} and OpenHA~\citep{wang2025openhaseriesopensourcehierarchical}, Qwen2-VL-7B-based vision--language--action models; and Open P2P~\citep{yue2026scalingbehaviorcloningimproves}, a 1.2B-parameter real-time gaming policy not trained on Minecraft. Each model is queried at the onset of the human response, using its supported inputs: preceding gameplay frames (ranging from 3 for JARVIS-VLA to 200, or 10~s, for Open P2P), corresponding recorded keyboard/mouse actions converted to the model's native format, and the transcribed instruction. Using the released inference implementations and default sampling settings, each query predicts a single 50~ms action, evaluated against the human's key presses during the same interval. We perform 10 runs per prompt and report $D_K$, the Hamming distance between the requested key and predicted pressed-key set, with 95\% confidence intervals obtained by bootstrapping over prompts and runs.

\begin{figure}[!t]
  \centering
  \includegraphics[width=\linewidth]{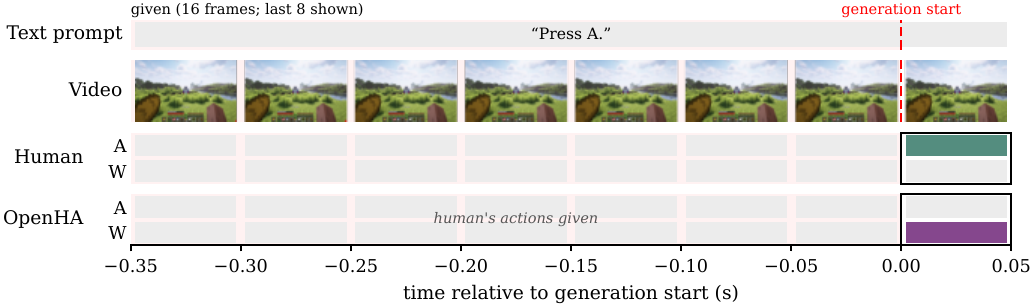}
  \caption{Example of one OpenHA run. Given the text prompt, the frames and the human's actions up to the generation start, it generates one 50 ms action}
\label{fig:experiment_example_openha}
\end{figure}

\Cref{fig:experiment_summary} summarizes the results, with per-prompt breakdowns in \Cref{fig:experiment_dk_per_prompt} and an example in \Cref{fig:experiment_example_openha}. All models struggled with these basic instructions. Only OpenHA and JARVIS-VLA produced the exact requested key, 17 in 100 for OpenHA and 1 in 100 for JARVIS-VLA; the other models never succeeded. Most predictions contained no key presses, nonempty predictions predominantly involved movement keys (W, Space, Ctrl), often unrelated to the instruction. Despite their demonstrated capabilities on complex Minecraft tasks, these agents struggle to follow even simple key-press instructions. Furthermore, none supports speech generation, preventing them from participating in the verbal social interactions central to PLAICraft.




\section{Discussion}\label{sec:discussion}

PLAICraft enables new lines of inquiry in embodied AI research. The inclusion of speech audio and the multiplayer social environments can support agents to incorporate real-time social cues, fostering collaborative play that prior single-player datasets cannot capture. The task-free, continual world and agent states also lend themselves to \textbf{continual learning} \citep{Kirkpatrick_2017} paradigms, where agents must develop and maintain long-horizon memory over days or weeks of game time while continually adapting to a non-stationary environment. The same long, rich trajectories, collected from the perspective of different players, can also enable systematic studies of \textbf{third-person learning} \citep{stadie2019thirdpersonimitationlearning}: training agents from logs of other players’ behaviour rather than from their own on-policy experience. This setting, with controllable data and compute budgets, supports quantitative investigation of \textbf{scaling laws} \citep{pearce2024scalinglawspretrainingagents} for embodied agents, relating training resources to control, communication, and social coordination capabilities. The vast collection of multimodal data also makes PLAICraft well-suited for \textbf{world model, 3D, and physics simulation} research. Since the initial release of the dataset last year, it has benefited various research in different areas, such as continual learning \citep{yoo2024lifelonglearningvideodiffusion}, imitation learning and diffusion models \citep{liu2026mirrorlearning}.

Our data collection platform is \textbf{application-agnostic}: because it logs raw keyboard/mouse and audio around a fixed display stream inside a controlled VM, we can easily replace Minecraft with any desktop application (e.g., web browser, productivity or creative tools) while preserving the same data schema and time-alignment guarantees. This allows us to extend beyond Minecraft agents to \textbf{general computer-use agents} \citep{xie2024osworldbenchmarkingmultimodalagents} that can operate a full computer interface at high temporal precision. The value of computer-use agents does not stop at information workflow. Remote teleoperators of warehouse and other robots are able to actuate complex robots and other cyberphysical infrastructure entirely through a standard computer interface, albeit with the added requirement of high time-precision ``dexterous'' keyboard and mouse actuation. Computer use agents that learn to mimic teleoperation control inputs to embodied robot agents have the potential to become de facto control policies of such robots.

\section{Conclusion}\label{sec:conclusion}

We introduced PLAICraft, a standardized data-collection platform and large-scale multimodal dataset. We also introduced an evaluation suite aimed at probing the important capabilities a human-like agent should have. PLAICraft is designed to open new avenues for research in social agents, continual and third-person learning, world-model, general computer-use and teleoperation-style agents.

\newpage
\clearpage
\section{LLM Usage Statement}
Large language models (LLMs) were used to assist with manuscript preparation and code development. Specifically, ChatGPT was used to refine author-written text, including grammatical corrections and improvements in clarity and readability. Claude was used to provide limited assistance with code implementation, debugging, and cleanup. Additionally, an LLM-based analysis pipeline was employed to characterize conversational interactions and label gameplay segments based on audio transcripts, using DeepSeek-V3.2-fast, as described in \Cref{sec3.3:social_interactions}. These automated annotations were used for dataset analysis and participant interaction assessment. All research motivations, scientific ideas, methodological designs, and experiments were independently developed and conducted by the authors. No LLMs were used beyond the applications disclosed above.
{
    \small
    \bibliographystyle{ieeenat_fullname}
    \bibliography{main}

@String(CVPR= {IEEE Conf. Comput. Vis. Pattern Recog.})

@String(ICCV= {Int. Conf. Comput. Vis.})

@String(IJCAI = {IJCAI})

@String(AAAI = {AAAI})

@String(CVPR  = {CVPR})

@String(ICCV  = {ICCV})

@misc{deepseekai2025deepseekv32pushingfrontieropen,
      title={DeepSeek-V3.2: Pushing the Frontier of Open Large Language Models}, 
      author={DeepSeek-AI and Aixin Liu and Aoxue Mei and Bangcai Lin and Bing Xue and Bingxuan Wang and Bingzheng Xu and Bochao Wu and Bowei Zhang and Chaofan Lin and Chen Dong and Chengda Lu and Chenggang Zhao and Chengqi Deng and Chenhao Xu and Chong Ruan and Damai Dai and Daya Guo and Dejian Yang and Deli Chen and Erhang Li and Fangqi Zhou and Fangyun Lin and Fucong Dai and Guangbo Hao and Guanting Chen and Guowei Li and H. Zhang and Hanwei Xu and Hao Li and Haofen Liang and Haoran Wei and Haowei Zhang and Haowen Luo and Haozhe Ji and Honghui Ding and Hongxuan Tang and Huanqi Cao and Huazuo Gao and Hui Qu and Hui Zeng and Jialiang Huang and Jiashi Li and Jiaxin Xu and Jiewen Hu and Jingchang Chen and Jingting Xiang and Jingyang Yuan and Jingyuan Cheng and Jinhua Zhu and Jun Ran and Junguang Jiang and Junjie Qiu and Junlong Li and Junxiao Song and Kai Dong and Kaige Gao and Kang Guan and Kexin Huang and Kexing Zhou and Kezhao Huang and Kuai Yu and Lean Wang and Lecong Zhang and Lei Wang and Liang Zhao and Liangsheng Yin and Lihua Guo and Lingxiao Luo and Linwang Ma and Litong Wang and Liyue Zhang and M. S. Di and M. Y Xu and Mingchuan Zhang and Minghua Zhang and Minghui Tang and Mingxu Zhou and Panpan Huang and Peixin Cong and Peiyi Wang and Qiancheng Wang and Qihao Zhu and Qingyang Li and Qinyu Chen and Qiushi Du and Ruiling Xu and Ruiqi Ge and Ruisong Zhang and Ruizhe Pan and Runji Wang and Runqiu Yin and Runxin Xu and Ruomeng Shen and Ruoyu Zhang and S. H. Liu and Shanghao Lu and Shangyan Zhou and Shanhuang Chen and Shaofei Cai and Shaoyuan Chen and Shengding Hu and Shengyu Liu and Shiqiang Hu and Shirong Ma and Shiyu Wang and Shuiping Yu and Shunfeng Zhou and Shuting Pan and Songyang Zhou and Tao Ni and Tao Yun and Tian Pei and Tian Ye and Tianyuan Yue and Wangding Zeng and Wen Liu and Wenfeng Liang and Wenjie Pang and Wenjing Luo and Wenjun Gao and Wentao Zhang and Xi Gao and Xiangwen Wang and Xiao Bi and Xiaodong Liu and Xiaohan Wang and Xiaokang Chen and Xiaokang Zhang and Xiaotao Nie and Xin Cheng and Xin Liu and Xin Xie and Xingchao Liu and Xingkai Yu and Xingyou Li and Xinyu Yang and Xinyuan Li and Xu Chen and Xuecheng Su and Xuehai Pan and Xuheng Lin and Xuwei Fu and Y. Q. Wang and Yang Zhang and Yanhong Xu and Yanru Ma and Yao Li and Yao Li and Yao Zhao and Yaofeng Sun and Yaohui Wang and Yi Qian and Yi Yu and Yichao Zhang and Yifan Ding and Yifan Shi and Yiliang Xiong and Ying He and Ying Zhou and Yinmin Zhong and Yishi Piao and Yisong Wang and Yixiao Chen and Yixuan Tan and Yixuan Wei and Yiyang Ma and Yiyuan Liu and Yonglun Yang and Yongqiang Guo and Yongtong Wu and Yu Wu and Yuan Cheng and Yuan Ou and Yuanfan Xu and Yuduan Wang and Yue Gong and Yuhan Wu and Yuheng Zou and Yukun Li and Yunfan Xiong and Yuxiang Luo and Yuxiang You and Yuxuan Liu and Yuyang Zhou and Z. F. Wu and Z. Z. Ren and Zehua Zhao and Zehui Ren and Zhangli Sha and Zhe Fu and Zhean Xu and Zhenda Xie and Zhengyan Zhang and Zhewen Hao and Zhibin Gou and Zhicheng Ma and Zhigang Yan and Zhihong Shao and Zhixian Huang and Zhiyu Wu and Zhuoshu Li and Zhuping Zhang and Zian Xu and Zihao Wang and Zihui Gu and Zijia Zhu and Zilin Li and Zipeng Zhang and Ziwei Xie and Ziyi Gao and Zizheng Pan and Zongqing Yao and Bei Feng and Hui Li and J. L. Cai and Jiaqi Ni and Lei Xu and Meng Li and Ning Tian and R. J. Chen and R. L. Jin and S. S. Li and Shuang Zhou and Tianyu Sun and X. Q. Li and Xiangyue Jin and Xiaojin Shen and Xiaosha Chen and Xinnan Song and Xinyi Zhou and Y. X. Zhu and Yanping Huang and Yaohui Li and Yi Zheng and Yuchen Zhu and Yunxian Ma and Zhen Huang and Zhipeng Xu and Zhongyu Zhang and Dongjie Ji and Jian Liang and Jianzhong Guo and Jin Chen and Leyi Xia and Miaojun Wang and Mingming Li and Peng Zhang and Ruyi Chen and Shangmian Sun and Shaoqing Wu and Shengfeng Ye and T. Wang and W. L. Xiao and Wei An and Xianzu Wang and Xiaowen Sun and Xiaoxiang Wang and Ying Tang and Yukun Zha and Zekai Zhang and Zhe Ju and Zhen Zhang and Zihua Qu},
      year={2025},
      eprint={2512.02556},
      archivePrefix={arXiv},
      primaryClass={cs.CL},
      url={https://arxiv.org/abs/2512.02556}, 
}

@misc{yue2026scalingbehaviorcloningimproves,
      title={Scaling Behavior Cloning Improves Causal Reasoning: An Open Model for Real-Time Video Game Playing}, 
      author={Yuguang Yue and Irakli Salia and Samuel Hunt and Chris Green and Wenzhe Shi and Jonathan J Hunt},
      year={2026},
      eprint={2601.04575},
      archivePrefix={arXiv},
      primaryClass={cs.AI},
      url={https://arxiv.org/abs/2601.04575}, 
}

@misc{wang2025openhaseriesopensourcehierarchical,
      title={OpenHA: A Series of Open-Source Hierarchical Agentic Models in Minecraft}, 
      author={Zihao Wang and Muyao Li and Kaichen He and Xiangyu Wang and Zhancun Mu and Anji Liu and Yitao Liang},
      year={2025},
      eprint={2509.13347},
      archivePrefix={arXiv},
      primaryClass={cs.AI},
      url={https://arxiv.org/abs/2509.13347}, 
}

@misc{li2025jarvisvlaposttraininglargescalevision,
      title={JARVIS-VLA: Post-Training Large-Scale Vision Language Models to Play Visual Games with Keyboards and Mouse}, 
      author={Muyao Li and Zihao Wang and Kaichen He and Xiaojian Ma and Yitao Liang},
      year={2025},
      eprint={2503.16365},
      archivePrefix={arXiv},
      primaryClass={cs.CV},
      url={https://arxiv.org/abs/2503.16365}, 
}

@misc{reddy2022dnsmosp835nonintrusiveperceptual,
      title={DNSMOS P.835: A Non-Intrusive Perceptual Objective Speech Quality Metric to Evaluate Noise Suppressors}, 
      author={Chandan K A Reddy and Vishak Gopal and Ross Cutler},
      year={2022},
      eprint={2110.01763},
      archivePrefix={arXiv},
      primaryClass={eess.AS},
      url={https://arxiv.org/abs/2110.01763}, 
}

@misc{sekoyan2025canary1bv2parakeettdt06bv3efficient,
      title={Canary-1B-v2 \& Parakeet-TDT-0.6B-v3: Efficient and High-Performance Models for Multilingual ASR and AST}, 
      author={Monica Sekoyan and Nithin Rao Koluguri and Nune Tadevosyan and Piotr Zelasko and Travis Bartley and Nikolay Karpov and Jagadeesh Balam and Boris Ginsburg},
      year={2025},
      eprint={2509.14128},
      archivePrefix={arXiv},
      primaryClass={cs.CL},
      url={https://arxiv.org/abs/2509.14128}, 
}

@misc{koluguri2021titanetneuralmodelspeaker,
      title={TitaNet: Neural Model for speaker representation with 1D Depth-wise separable convolutions and global context}, 
      author={Nithin Rao Koluguri and Taejin Park and Boris Ginsburg},
      year={2021},
      eprint={2110.04410},
      archivePrefix={arXiv},
      primaryClass={eess.AS},
      url={https://arxiv.org/abs/2110.04410}, 
}

@misc{park2025sortformernovelapproachpermutationresolved,
      title={Sortformer: A Novel Approach for Permutation-Resolved Speaker Supervision in Speech-to-Text Systems}, 
      author={Taejin Park and Ivan Medennikov and Kunal Dhawan and Weiqing Wang and He Huang and Nithin Rao Koluguri and Krishna C. Puvvada and Jagadeesh Balam and Boris Ginsburg},
      year={2025},
      eprint={2409.06656},
      archivePrefix={arXiv},
      primaryClass={eess.AS},
      url={https://arxiv.org/abs/2409.06656}, 
}

@misc{fu2026rethinkingtrainingtargetsarchitectures,
      title={Rethinking Training Targets, Architectures and Data Quality for Universal Speech Enhancement}, 
      author={Szu-Wei Fu and Rong Chao and Xuesong Yang and Sung-Feng Huang and Ryandhimas E. Zezario and Rauf Nasretdinov and Ante Jukić and Yu Tsao and Yu-Chiang Frank Wang},
      year={2026},
      eprint={2603.02641},
      archivePrefix={arXiv},
      primaryClass={cs.SD},
      url={https://arxiv.org/abs/2603.02641}, 
}

@misc{liu2026mirrorlearning,
      title={Mirror Learning}, 
      author={Yunpeng Liu and Matthew Niedoba and Oluwanifemi A. Adekanye and Jason Yoo and Yingchen He and Berend Zwartsenberg and Frank Wood},
      year={2026},
      eprint={2607.28737},
      archivePrefix={arXiv},
      primaryClass={cs.LG},
      url={https://arxiv.org/abs/2607.28737}, 
}

@misc{xie2024osworldbenchmarkingmultimodalagents,
      title={OSWorld: Benchmarking Multimodal Agents for Open-Ended Tasks in Real Computer Environments}, 
      author={Tianbao Xie and Danyang Zhang and Jixuan Chen and Xiaochuan Li and Siheng Zhao and Ruisheng Cao and Toh Jing Hua and Zhoujun Cheng and Dongchan Shin and Fangyu Lei and Yitao Liu and Yiheng Xu and Shuyan Zhou and Silvio Savarese and Caiming Xiong and Victor Zhong and Tao Yu},
      year={2024},
      eprint={2404.07972},
      archivePrefix={arXiv},
      primaryClass={cs.AI},
      url={https://arxiv.org/abs/2404.07972}, 
}

@misc{pearce2024scalinglawspretrainingagents,
      title={Scaling Laws for Pre-training Agents and World Models}, 
      author={Tim Pearce and Tabish Rashid and Dave Bignell and Raluca Georgescu and Sam Devlin and Katja Hofmann},
      year={2024},
      eprint={2411.04434},
      archivePrefix={arXiv},
      primaryClass={cs.LG},
      url={https://arxiv.org/abs/2411.04434}, 
}

@misc{li2017learningforgetting,
      title={Learning without Forgetting}, 
      author={Zhizhong Li and Derek Hoiem},
      year={2017},
      eprint={1606.09282},
      archivePrefix={arXiv},
      primaryClass={cs.CV},
      url={https://arxiv.org/abs/1606.09282}, 
}

@article{Kirkpatrick_2017,
   title={Overcoming catastrophic forgetting in neural networks},
   volume={114},
   ISSN={1091-6490},
   url={http://dx.doi.org/10.1073/pnas.1611835114},
   DOI={10.1073/pnas.1611835114},
   number={13},
   journal={Proceedings of the National Academy of Sciences},
   publisher={Proceedings of the National Academy of Sciences},
   author={Kirkpatrick, James and Pascanu, Razvan and Rabinowitz, Neil and Veness, Joel and Desjardins, Guillaume and Rusu, Andrei A. and Milan, Kieran and Quan, John and Ramalho, Tiago and Grabska-Barwinska, Agnieszka and Hassabis, Demis and Clopath, Claudia and Kumaran, Dharshan and Hadsell, Raia},
   year={2017},
   month=Mar, pages={3521–3526} }

@misc{stadie2019thirdpersonimitationlearning,
      title={Third-Person Imitation Learning}, 
      author={Bradly C. Stadie and Pieter Abbeel and Ilya Sutskever},
      year={2019},
      eprint={1703.01703},
      archivePrefix={arXiv},
      primaryClass={cs.LG},
      url={https://arxiv.org/abs/1703.01703}, 
}

@misc{wang2024comprehensivesurveycontinuallearning,
      title={A Comprehensive Survey of Continual Learning: Theory, Method and Application}, 
      author={Liyuan Wang and Xingxing Zhang and Hang Su and Jun Zhu},
      year={2024},
      eprint={2302.00487},
      archivePrefix={arXiv},
      primaryClass={cs.LG},
      url={https://arxiv.org/abs/2302.00487}, 
}

@article{brooks1991intelligence,
  title={Intelligence without representation},
  author={Brooks, Rodney A},
  journal={Artificial intelligence},
  volume={47},
  number={1-3},
  pages={139--159},
  year={1991},
  publisher={Elsevier}
}

@inproceedings{
milani2023bedd,
title={{BEDD}: The Mine{RL} {BASALT} Evaluation and Demonstrations Dataset for Training and Benchmarking Agents that Solve Fuzzy Tasks},
author={Stephanie Milani and Anssi Kanervisto and Karolis Ramanauskas and Sander V Schulhoff and Brandon Houghton and Rohin Shah},
booktitle={Thirty-seventh Conference on Neural Information Processing Systems Datasets and Benchmarks Track},
year={2023},
url={https://openreview.net/forum?id=D1MOK2t2t2}
}

@misc{shah2021minerl,
      title={The MineRL BASALT Competition on Learning from Human Feedback}, 
      author={Rohin Shah and Cody Wild and Steven H. Wang and Neel Alex and Brandon Houghton and William Guss and Sharada Mohanty and Anssi Kanervisto and Stephanie Milani and Nicholay Topin and Pieter Abbeel and Stuart Russell and Anca Dragan},
      year={2021},
      eprint={2107.01969},
      archivePrefix={arXiv},
      primaryClass={cs.LG}
}

@inproceedings{10.5555/3061053.3061259,
author = {Johnson, Matthew and Hofmann, Katja and Hutton, Tim and Bignell, David},
title = {The Malmo platform for artificial intelligence experimentation},
year = {2016},
isbn = {9781577357704},
publisher = {AAAI Press},
booktitle = {Proceedings of the Twenty-Fifth International Joint Conference on Artificial Intelligence},
pages = {4246–4247},
numpages = {2},
location = {New York, New York, USA},
series = {IJCAI'16}
}

@misc{fan2022minedojo,
      title={MineDojo: Building Open-Ended Embodied Agents with Internet-Scale Knowledge}, 
      author={Linxi Fan and Guanzhi Wang and Yunfan Jiang and Ajay Mandlekar and Yuncong Yang and Haoyi Zhu and Andrew Tang and De-An Huang and Yuke Zhu and Anima Anandkumar},
      year={2022},
      eprint={2206.08853},
      archivePrefix={arXiv},
      primaryClass={cs.LG}
}

@misc{guss2019minerl,
      title={MineRL: A Large-Scale Dataset of Minecraft Demonstrations}, 
      author={William H. Guss and Brandon Houghton and Nicholay Topin and Phillip Wang and Cayden Codel and Manuela Veloso and Ruslan Salakhutdinov},
      year={2019},
      eprint={1907.13440},
      archivePrefix={arXiv},
      primaryClass={cs.LG}
}

@article{Zhou2024MineDreamerLT,
  title={MineDreamer: Learning to Follow Instructions via Chain-of-Imagination for Simulated-World Control},
  author={Enshen Zhou and Yiran Qin and Zhen-fei Yin and Yuzhou Huang and Ruimao Zhang and Lu Sheng and Yu Qiao and Jing Shao},
  journal={ArXiv},
  year={2024},
  volume={abs/2403.12037},
  url={https://api.semanticscholar.org/CorpusID:268532481}
}

@misc{lifshitz2024steve1,
      title={STEVE-1: A Generative Model for Text-to-Behavior in Minecraft}, 
      author={Shalev Lifshitz and Keiran Paster and Harris Chan and Jimmy Ba and Sheila McIlraith},
      year={2024},
      eprint={2306.00937},
      archivePrefix={arXiv},
      primaryClass={cs.AI}
}

@article{devin2024,
  title={Devin: The first AI software engineer},
  author={Cognition AI},
  note={\url{https://www.cognition-labs.com/}},
  year={2024}
}

@article{das2019tacl,
  title={Building Modular and Compositional Agents with the CraftAssist Framework},
  author={Das, Abhishek and et al.},
  journal={Transactions of the ACL},
  year={2019}
}

@inproceedings{savva2019habitat,
  title={Habitat: A Platform for Embodied AI Research},
  author={Savva, Manolis and et al.},
  booktitle={ICCV},
  year={2019}
}

@inproceedings{deitke2020robothor,
  title={RoboTHOR: An Open Simulation-to-Real Embodied AI Platform},
  author={Deitke, Mitchell and et al.},
  booktitle={CVPR Workshops},
  year={2020}
}

@inproceedings{gan2021threedworld,
  title={ThreeDWorld: A Platform for Interactive Multi-Modal Physical Simulation},
  author={Gan, Chuang and et al.},
  booktitle={NeurIPS},
  year={2021}
}

@misc{bain2023whisperxtimeaccuratespeechtranscription,
      title={WhisperX: Time-Accurate Speech Transcription of Long-Form Audio}, 
      author={Max Bain and Jaesung Huh and Tengda Han and Andrew Zisserman},
      year={2023},
      eprint={2303.00747},
      archivePrefix={arXiv},
      primaryClass={cs.SD},
      url={https://arxiv.org/abs/2303.00747}, 
}

@article{Voudouris2022-VOUDHC,
	author = {Konstantinos Voudouris and Matthew Crosby and Benjamin Beyret and Jos\'e Hern\'{a}ndez{-}Orallo and Murray Shanahan and Marta Halina and Lucy G. Cheke},
	doi = {10.3389/fpsyg.2022.711821},
	journal = {Frontiers in Psychology},
	title = {Direct Human-Ai Comparison in the Animal-Ai Environment},
	volume = {13},
	year = {2022}
}

@misc{gandhi2022babyintuitionsbenchmarkbib,
      title={Baby Intuitions Benchmark (BIB): Discerning the goals, preferences, and actions of others}, 
      author={Kanishk Gandhi and Gala Stojnic and Brenden M. Lake and Moira R. Dillon},
      year={2022},
      eprint={2102.11938},
      archivePrefix={arXiv},
      primaryClass={cs.AI},
      url={https://arxiv.org/abs/2102.11938}, 
}

@misc{lake2016buildingmachineslearnthink,
      title={Building Machines That Learn and Think Like People}, 
      author={Brenden M. Lake and Tomer D. Ullman and Joshua B. Tenenbaum and Samuel J. Gershman},
      year={2016},
      eprint={1604.00289},
      archivePrefix={arXiv},
      primaryClass={cs.AI},
      url={https://arxiv.org/abs/1604.00289}, 
}

@misc{yoo2024lifelonglearningvideodiffusion,
      title={Lifelong Learning of Video Diffusion Models From a Single Video Stream}, 
      author={Jason Yoo and Yingchen He and Saeid Naderiparizi and Dylan Green and Gido M. van de Ven and Geoff Pleiss and Frank Wood},
      year={2024},
      eprint={2406.04814},
      archivePrefix={arXiv},
      primaryClass={cs.CV},
      url={https://arxiv.org/abs/2406.04814}, 
}

@misc{wagner2024crisperwhisperaccuratetimestampsverbatim,
      title={CrisperWhisper: Accurate Timestamps on Verbatim Speech Transcriptions}, 
      author={Laurin Wagner and Bernhard Thallinger and Mario Zusag},
      year={2024},
      eprint={2408.16589},
      archivePrefix={arXiv},
      primaryClass={cs.LG},
      url={https://arxiv.org/abs/2408.16589}, 
}

@article{schneider2018cattell,
  title={The Cattell-Horn-Carroll theory of cognitive abilities},
  author={Schneider, W Joel and McGrew, Kevin S},
  journal={Contemporary intellectual assessment: Theories, tests, and issues},
  volume={733},
  pages={163},
  year={2018}
}

@book{carroll1993human,
  title={Human cognitive abilities: A survey of factor-analytic studies},
  author={Carroll, John Bissell},
  number={1},
  year={1993},
  publisher={Cambridge university press}
}

@misc{anthropic2024claude,
  author       = {Anthropic},
  title        = {Claude Code Overview},
  year         = {2024},
  howpublished = {\url{https://docs.anthropic.com/en/docs/claude-code/overview}},
  note         = {Accessed: 2025-05-15}
}

@misc{zurn2024wayvescenes101,
  author       = {Jannik Zürn and Paul Gladkov and Sofía Dudas and Fergal Cotter and Sofi Toteva and Jamie Shotton and Vasiliki Simaiaki and Nikhil Mohan},
  title        = {WayveScenes101: A Dataset and Benchmark for Novel View Synthesis in Autonomous Driving},
  year         = {2024},
  howpublished = {\url{https://wayve.ai/science/wayvescenes101/}},
  note         = {Accessed: 2025-05-15}
}

@misc{waymo2025e2e,
  author       = {Waymo},
  title        = {2025 Waymo Open Dataset Challenges: Vision-based End-to-End Driving},
  year         = {2025},
  howpublished = {\url{https://waymo.com/open/challenges/2025/e2e-driving/}},
  note         = {Accessed: 2025-05-15}
}

@misc{baker2022videopretrainingvptlearning,
      title={Video PreTraining (VPT): Learning to Act by Watching Unlabeled Online Videos}, 
      author={Bowen Baker and Ilge Akkaya and Peter Zhokhov and Joost Huizinga and Jie Tang and Adrien Ecoffet and Brandon Houghton and Raul Sampedro and Jeff Clune},
      year={2022},
      eprint={2206.11795},
      archivePrefix={arXiv},
      primaryClass={cs.LG},
      url={https://arxiv.org/abs/2206.11795}, 
}

@inproceedings{deng2023mind2web,
  title     = {Mind2Web: Towards a Generalist Agent for the Web},
  author    = {Deng, Xiang and others},
  booktitle = {Advances in Neural Information Processing Systems (NeurIPS), Datasets and Benchmarks Track},
  year      = {2023}
}

@inproceedings{sun2020waymo,
  title     = {Scalability in Perception for Autonomous Driving: Waymo Open Dataset},
  author    = {Sun, Pei and Kretzschmar, Henrik and Dotiwalla, Xerxes and Chouard, Aurelien and Patnaik, Vijaysai and Tsui, Paul and Guo, James and Zhou, Yin and Chai, Yuning and Caine, Benjamin and Vasudevan, Vijay and Han, Wei and Ngiam, Jiquan and Zhao, Hang and Timofeev, Aleksei and Ettinger, Scott and Krivokon, Maxim and Gao, Amy and Joshi, Aditya and Zhang, Yu and Shlens, Jonathon and Chen, Zhifeng and Anguelov, Dragomir},
  booktitle = {Proceedings of the IEEE/CVF Conference on Computer Vision and Pattern Recognition (CVPR)},
  year      = {2020},
  pages     = {2446--2454}
}

@misc{valevski2025diffusionmodelsrealtimegame,
      title={Diffusion Models Are Real-Time Game Engines}, 
      author={Dani Valevski and Yaniv Leviathan and Moab Arar and Shlomi Fruchter},
      year={2025},
      eprint={2408.14837},
      archivePrefix={arXiv},
      primaryClass={cs.LG},
      url={https://arxiv.org/abs/2408.14837}, 
}
}

\clearpage
\setcounter{page}{1}

\appendix
\section{Appendix}

\subsection{Ethical Considerations}
\label{sec:ethics}
This research ensures the safety and privacy of participants through the guidance of our university's office of research ethics. Study procedures are reviewed routinely by the Behavioural Research Ethics Board (BREB); and have been found acceptable on ethical grounds for research involving human subjects, as well as deemed minimal risk to participants. Participants’ data is collected with their consent, or that of their guardian, and any reasonably identifiable aspects of this data are anonymized systematically.

\subsection{Societal Impacts}
\label{sec:societal_impacts}

Combining embodied AI with immersive game environments has the potential to amplify both the benefits and risks of artificial intelligence. On the upside, interactive gameplay environments like Minecraft provide a uniquely rich substrate for training and evaluating agents with social, linguistic, and perceptual competencies. Agents trained in such environments could support educational gameplay, serve as in-game tutors, or scaffold more collaborative and engaging human experiences. As embodied agents become more fluent in social cues—especially through real-time speech and interaction—they may also help simulate realistic training scenarios or serve therapeutic and accessibility functions.

At the same time, advances in speech-enabled embodied agents may blur the line between artificial and human players. Players may not always be able to distinguish between synthetic and human teammates or adversaries. While this ambiguity can enhance immersion, it may also raise concerns about deception, manipulation, or the erosion of trust. Agents that convincingly imitate social behaviours could influence players in subtle ways, especially younger users. There is a risk that such agents could be used to simulate companionship without reciprocity, or to manipulate attention and emotional responses for commercial ends.

The collection of large-scale, multimodal human gameplay data also raises important questions about consent, privacy, and downstream use. We have undergone an institutional privacy review and obtained consent for data release, and have designed our infrastructure to preserve anonymity and prevent the recording of persistent identity information. Nonetheless, the richness of multimodal recordings—particularly audio—makes full anonymization difficult. Developers and researchers using such data should exercise caution in avoiding reidentification risks and consider the broader implications of using data collected in seemingly informal or playful contexts to train powerful AI systems.

Finally, embodied AI in gaming environments may raise labour and cultural concerns. As agents become capable of performing in-game roles—e.g., assistants, performers, or economic participants—they could displace human workers in virtual economies, reshape online social norms, or saturate game environments with synthetic interactions. While our work is aimed at research and foundational exploration, we believe that careful governance, transparency, and participatory design will be essential as embodied agents become more integrated into human play and work environments.

\subsection{Minecraft Configurations} \label{sec: Minecraft configuration}
\textbf{Client Side:} We retained most features of vanilla Java Minecraft. All three worlds (the Overworld, the Nether, and the End) are accessible to players. Every player ran the same client instance: Minecraft Java Edition 1.21.4 on NeoForge 21.4.136, in fullscreen, first-person view. The field of view was the default 70$^\circ$ and mouse sensitivity was the default 100\%. Raw mouse input, auto-jump and smooth camera were off. Sprint and crouch were hold-to-activate, and all key bindings were vanilla defaults (WASD to move, Space to jump, Left Shift to crouch, Left Ctrl to sprint, 1--9 for the hotbar, E for inventory). Rendering used Fancy graphics, view bobbing, brightness at 80\%, an automatic GUI scale, decreased particles, and render and simulation distances of 8 chunks. The frame rate was capped at 60 FPS with V-Sync. Background music was muted. Players talked to each other through the Simple Voice Chat mod, which plays speech positionally within 48 blocks and transmits on voice activation rather than push-to-talk.

We restricted access to the in-game settings menu so that players could neither exit the game nor change the configuration above. If a player leaves the Minecraft window, the instance is shut down and the data recorded up to that point is uploaded, which ensures that all captured data consists solely of gameplay. We also implemented a mod that captures players' keyboard and mouse actions with more precise timestamps than system-level logging. This brings the recorded actions closer to the Minecraft client tick and so better aligns cause and effect with the rendered frames. To encourage meaningful interaction, we built a live monitoring pipeline that uses WhisperX and large language models to analyze the audio each player speaks and hears. We use its output to measure the quality of interaction between players and reward them accordingly. Finally, we incorporated a range of third-party client-side mods and plugins to improve performance, visuals, and playability, as shown in \Cref{tab:client-side-plugins}. These include Sodium and Iris with the Complementary Unbound (r5.5.1) shader pack.\\
\textbf{Server Side:} First, we developed a custom plugin, AutoJoin, to manage player states. Player information, including inventory, level, and location, is stored in our database. When a player joins, a random Minecraft license is assigned, and the plugin restores the player's saved state and teleports them to their previous location. Upon leaving, the player's state is saved back to the database, ensuring a seamless experience while enabling efficient reuse of Minecraft licenses.

To encourage social interactions, we introduced three additional worlds beyond the default Minecraft dimensions. Players can access a bounded lobby world using the "/minigames" command, providing a safe space to meet and interact. From the lobby, players can enter two minigame worlds featuring BedWars and Murder Mystery, both intended to promote participation and player interactions.

Additionally, we integrated various third-party server-side plugins, as listed in \Cref{tab:server-side-plugins}.

\begin{table*}[ht]
  \centering
  \begin{tabular}{p{0.30\textwidth} p{0.65\textwidth}}
    \hline
    \textbf{Name} & \textbf{Description} \\
    \hline
    Complementary Shader Pack & Significantly enhances visual detail and overall graphical quality. \\
    Simple Voice Chat & Proximity-based voice chat enabling real-time communication. \\
    Canary & Performance optimization mod for general client-side speedups. \\
    Clumps & Reduces item-entity lag by grouping dropped items into stacks. \\
    EntityCulling & Skips rendering of off-screen entities to improve FPS. \\
    Fastload-Reforged & Accelerates startup and world-loading times. \\
    FerriteCore & Minimizes memory usage by stripping unused block states. \\
    ModernFix & Addresses rendering performance issues and chunk-loading bugs. \\
    Oculus & Optimizes block-update and rendering code paths for better FPS. \\
    Redirector & Replaces generic implementations with faster runtime alternatives. \\
    Rubidium & Speeds up core rendering operations via low-level optimizations. \\
    Saturn & Bundles multiple performance tweaks for smoother gameplay. \\
    SmoothBoot & Optimizes boot routines for faster game initialization. \\
    Starlight & Overhauls the lighting engine for much faster chunk light updates. \\
    BetterF3 & Enhanced debug overlay with customizable performance stats. \\
    Cloth-Config & Provides an in-game, user-friendly configuration GUI. \\
    Entity-Model-Features & Adds extra model customization hooks for entities. \\
    Entity-Texture-Features & Enables per-entity texture overrides and enhancements. \\
    NotEnoughAnimation & Introduces additional character animations for more realism. \\
    SkinLayer3D & Renders player skin layers in true 3D for added depth. \\
    \hline
  \end{tabular}
  \vspace{0.3em}
  \caption{Client-Side Mods \& Plugins}
  \label{tab:client-side-plugins}
\end{table*}

\begin{table*}[tbp]
  \centering
  \begin{tabular}{p{0.30\textwidth} p{0.65\textwidth}}
    \hline
    \textbf{Name} & \textbf{Description} \\
    \hline
    AFK-Kick & Prevents idle players by kicking after inactivity. \\
    BetterStructures & Adds new, interesting auto-generated structures. \\
    Better-RTP & Enhances random teleportation experience. \\
    ChatFilter & Filters and blocks offensive chat content. \\
    Screaming Bedwars & Implements Bedwars minigame mechanics. \\
    Chunky & Preloads chunks to reduce lag spikes. \\
    CoreProtect & Logs and rolls back griefing actions. \\
    DecentHolograms & Displays floating holograms without dependencies. \\
    DiscordSRV & Bridges Discord and Minecraft chat. \\
    EssentialsX & Provides core server management commands. \\
    GravesX & Spawns lootable death chests at player death. \\
    HideNametag & Removes player nametags for clean visuals. \\
    InvisibleItemFrames & Hides item frame visuals in-game. \\
    Multiverse-Core & Manages multiple worlds on one server. \\
    PlaceholderAPI & Provides a uniform placeholder system for other plugins. \\
    ProtocolLib & Allows manipulation of Minecraft network packets. \\
    SuperVanish & Enables admins to be invisible to players. \\
    \hline
  \end{tabular}
  \vspace{0.3em}
  \caption{Server-Side Plugins}
  \label{tab:server-side-plugins}
\end{table*}

Collectively, these modifications create a natural, intuitive, and enjoyable environment where participants can play Minecraft as they normally would, while also providing us with standardized, time-synchronized, multi-modal gameplay data.

\subsection{Additional Participants Demographics}\label{sec:participants_demographics}
\begin{figure*}[!htbp]
  \centering
  \includegraphics[width=\textwidth]{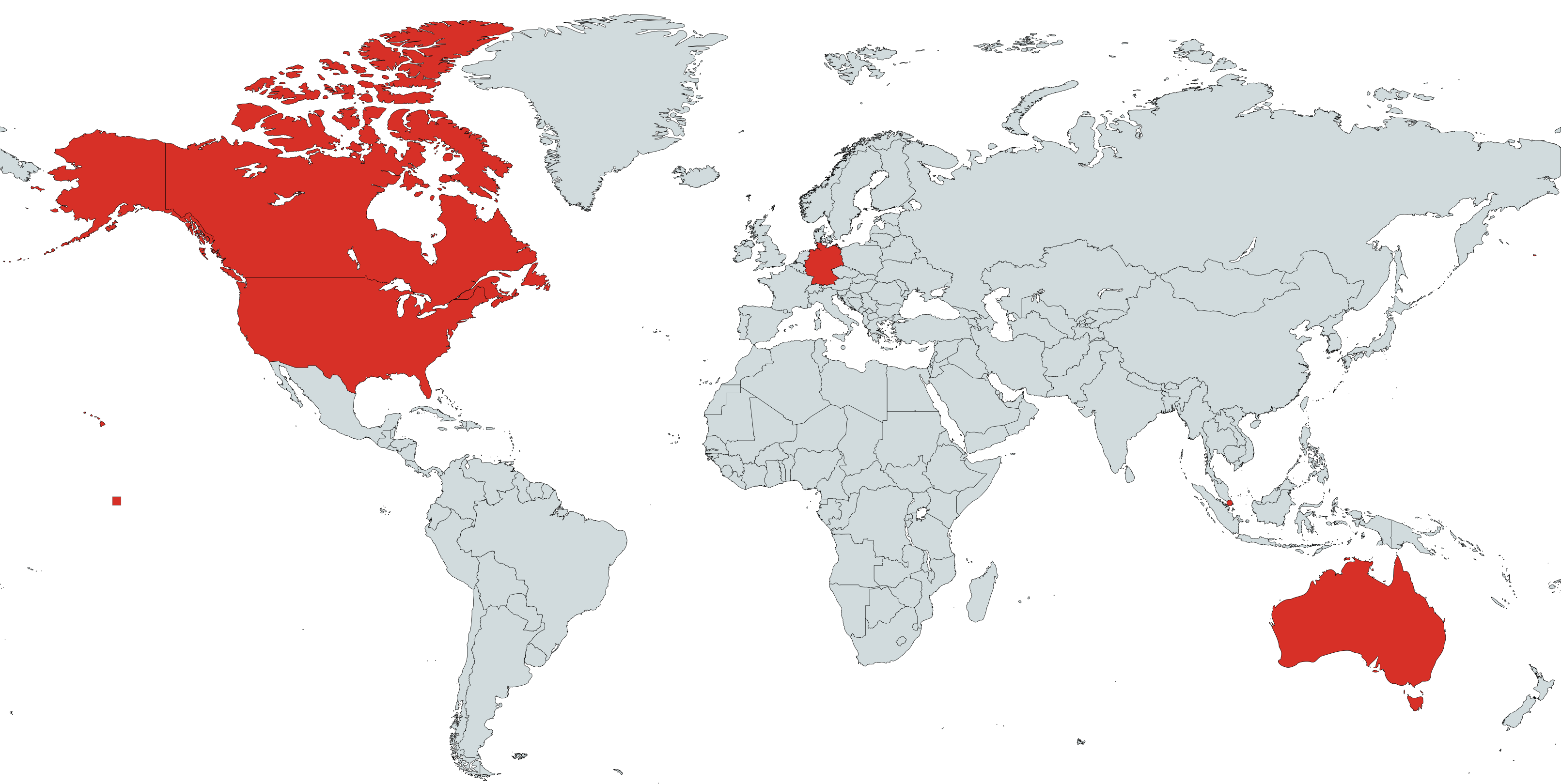}
  \caption{
    Our platform's deployment regions. Participants are not restricted to these countries as countries/regions close to these areas can still get a low latency when playing on our server. 
  }
  \label{fig:aws_deployment_region}
\end{figure*}

Our participants consist of Minecraft players across the world. To lower the overall latency players will experience, we deploy our EC2 instances in various locations across the world. Figure \ref{fig:aws_deployment_region} lists out our deployment regions. Players near these areas will have a low latency when playing through our streaming. Players further away from these regions are not restricted to access our platform, but their gaming experience might suffer from lagginess due to higher latency. 

We do not hire specific participants to conduct our data collection process, as it is financially impractical, nor do we restrict anyone from participating. Instead, we focus on promoting our platform as a research project where participants can play free Minecraft with beautiful shaders on any pc device and make it available to everyone. We have done a series of promotions, including local poster distribution, Google Ads, YouTube and TikTok content creation. As a result, we get a constant stream of old and new participants joining our server every day. By the time of the publication, we had reached 18,117 participants.

Our players come from three main sources: (1) \textbf{Minecraft fans reached through our outreach.} We performed an extensive media outreach and promotions, including TikTok, YouTube content creation, collaborating with famous Minecraft YouTubers, poster distributions and developed a blog website. Early growth of this category was slow, but once our promotions reached some communities, such as a school or Discord group, sign-ups surged. These players usually log the most hours and contribute the most advanced and valuable data. (2) \textbf{Members of our development team.} A few of our teammates who love Minecraft have added several hundred hours of play. They often help organize groups and build shared bases to boost social interactions on the server. Their contribution rate is significantly smaller than the other two groups. (3) \textbf{Undergraduate students earning extra credit.} We partnered with universities' courses that offered credit for playing a set number of hours. The participation rate was high at one point, though most of these students were Minecraft beginners and did not play for long. Since the program ended, the majority of this group has stopped contributing as well.

\begin{table*}[tbp]
          \centering
          \begin{tabular}{p{0.35\textwidth} p{0.65\textwidth}}
            \hline
            \textbf{Label} & \textbf{Question} \\
            \hline
            Interactive & Is this segment interactive? \\
            Minecraft-related & Is this segment related to Minecraft gameplay? \\
            Offensive & Does this segment contain offensive or inappropriate language? \\
            Crafting & Are the players building or crafting something together? \\
            Exploring & Are the players actively navigating the Minecraft world? \\
            Teaching & Is one player helping another learn something about the game? \\
            Fighting & Are the players involved in or discussing combat?\\
            Game Action Quality & Is this segment a high-quality example of a gameplay interaction? \\
            \hline
            \end{tabular}
            \vspace{0.3em}
            \caption{Examples of Binary Labeling Questions}
          \label{tab:labels_questions}
\end{table*}

\subsection{Content Moderation}
To protect dataset integrity and community health, an active moderation team oversees the game server around the clock. Supported by automated plugins and help from veteran players, moderators can promptly detect, kick, and ban misbehaving users on sight. Because every gameplay action and chat message is recorded and logged, any incidents missed in real time can be traced and addressed retrospectively. These safeguards preserve a wholesome community atmosphere, sustain growth, and limit contamination of the collected data.

\subsection{Recording Software}
\label{sec:Recording Software}
\textbf{Video \& Audio} Video recordings are captured using Open Broadcaster Software (OBS) running on the EC2 instance. Videos are recorded at a constant frame rate of 30 FPS with a resolution of 1280x720, encoded using the H.264 codec. OBS is specifically configured to capture only the Minecraft application window in full-screen mode, ensuring that all recorded frames pertain solely to gameplay. Audio streams are simultaneously recorded by OBS, separately capturing user microphone input and game audio output. Each audio stream is recorded at a constant sampling rate of 48 kHz in mono format, encoded with the AAC codec. We take the exact timestamp when OBS’s ffmpeg muxer began writing the file as the start timestamp, and align all modalities to it.

\textbf{Mouse \& Keyboard} Mouse and keyboard inputs are precisely captured using a custom-developed Forge mod integrated directly into the client-side Minecraft application. This method provides superior timing accuracy by recording when the game server actually processes input events, unlike traditional system-level logs. Input events for both mouse and keyboard are recorded on an event-driven basis with millisecond precision, maintaining a maximum polling rate of 100 Hz, resulting in a minimum interval of 10 ms between consecutive data points. Additionally, comprehensive system-level logging using pyxhook is implemented as a secondary data collection measure. These system-level logs, although less precise in timestamp accuracy, serve primarily to validate the integrity and completeness of the mod-generated input data.

\subsection{Data Pre-processing}\label{sec: Data Pre-processing}
We conducted a systematic and rigorous pre-processing pipeline on the raw data, transforming it into a structured and accessible format for public release. 

\subsubsection{Video}
Video data were extracted from the original MKV format and converted into MP4 files with a standardized frame rate of 30 FPS. We trim the start of the video to align with the player's first keyboard clicks, and we trim off the end to match when the player quits the game. This way, we make sure the content in the video is mostly related to the actual gameplay. 

\subsubsection{Audio}
Each session contains two separately recorded audio tracks: the player's microphone input (\texttt{audio\_speak}) and the audio heard by the player (\texttt{audio\_hear}), including voice chat from other players. We process \texttt{audio\_speak} to isolate the player's speech and improve audio quality. The raw 48~kHz microphone recording is first resampled to 24~kHz and enhanced using NVIDIA RE-USE for noise and reverberation suppression, operating on 5~s segments with 50\% overlap. To distinguish the player's voice from other speakers captured by the microphone, we apply NVIDIA Streaming Sortformer (v2.1) for speaker diarization, supporting up to four speakers with 80~ms frames and an activity threshold of 0.5. We then extract TitaNet-Large speaker embeddings and establish a consistent voiceprint for each player across all their sessions. Specifically, speaker embeddings are clustered using average-linkage cosine distance, and the cluster containing the most speech is assigned to the player. Its duration-weighted mean embedding serves as the player's voiceprint, against which all diarized speakers are verified using a cosine-similarity threshold of 0.7. We retain only frames in which the player is the sole active speaker, thereby excluding silence, overlapping speech, and speech from other individuals. All excluded regions are silenced, with 10~ms raised-cosine fades applied at segment boundaries.

The retained speech segments are subsequently grouped into utterances. Consecutive segments are merged when their separation is less than 2~s, the subsequent segment is shorter than 3~s, and the resulting utterance does not exceed 28~s. Each utterance is resampled to 16~kHz and transcribed using NVIDIA Parakeet-TDT-0.6B-v3, with word-level timestamps aligned to the session timeline. We further filter utterances based on language and acoustic quality, discarding those containing non-ASCII transcript characters, matching a manually reviewed list of non-English utterances, or exhibiting a DNSMOS P.835 overall quality score (OVRL) of 2.0 or lower for utterances lasting at least 3~s. DNSMOS scores are averaged over 9.01~s windows. Both the audio and transcripts of discarded utterances are removed.

To capture non-verbal vocalizations omitted by conventional speech recognition, we additionally apply CrisperWhisper~2.0~\citep{wagner2024crisperwhisperaccuratetimestampsverbatim}, conditioned on the Parakeet transcript. It processes windows of up to 30~s and 70 words, preserving the original transcription and timestamps while inserting filled pauses and non-verbal annotations from a predefined set of 15 tags, including \texttt{[UM]}, \texttt{[UH]}, \texttt{[laughter]}, \texttt{[breath]}, \texttt{[cough]}, and \texttt{[sigh]}. 

In contrast, \texttt{audio\_hear} is preserved in its original 48~kHz format without enhancement or filtering to maintain the auditory experience of the player. Its corresponding transcript is stored in the session database with word-level timestamps at millisecond precision.

\subsubsection{Mouse \& Keyboard}
Mouse data preprocessing involved segregating mouse click events from mouse movement trajectories. Mouse movement data were aggregated into fixed temporal bins of 100 ms, each represented by arrays of dimensions (2, 10), corresponding to relative x and y movements across timestamps aligned to a maximum polling frequency of 100 Hz. Each data point was assigned to the nearest bin, and intervals lacking data were zero-padded. Mouse click events, analogous in structure to keyboard event data, were similarly binned into arrays of shape (79, 10), where 79 dimensions represent 76 distinct keyboard keys in addition to the left, right, and middle mouse buttons. This consistent binning procedure facilitated unified storage of both mouse movement and combined mouse-keyboard event data in an organized database, enabling flexible data querying and analysis. 

Raw mouse movement contains cursor jumps that the player did not make, which we removed before binning. Our logging mod reads the cursor position every 10~ms and records the change since the previous reading. Whenever a screen such as the inventory or chat opens or closes, Minecraft moves the cursor to the centre of the window, so the reading in which this happens adds the jump to the centre on top of the player's real movement. Whenever a reading's position lands exactly on a centre pixel and the previous reading's does not, we replaced that reading's movement with its offset from the centre pixel; all other readings were kept unchanged. We checked the correction against the server's log of camera yaw, which records exactly how far the camera turned.

We also excluded intervals in which the mouse data does not reflect the player's hand movement at all. Players connected through a remote desktop (NICE DCV). When the game window lost focus, the remote desktop could switch from relative to absolute pointer mode until the player clicked back into the game. During these intervals, each 10~ms reading records a multiple of the pointer's screen position instead of its movement, throwing the camera around (usually until it points straight down). We detected these readings by matching each mouse step against the remote desktop's own pointer log. These intervals cover 0.79\% of play and will be masked out for any future training. 

\subsection{Automatic Data Annotation}
\label{sec:automatic_data_annotation}

To identify subsets of highly interactive data, particularly relevant for Embodied AI, we developed a methodology leveraging audio transcribed by WhisperX ~\citep{bain2023whisperxtimeaccuratespeechtranscription}, and the capabilities of large language models (LLMs). We first defined highly interactive data as continuous, game-related interactions between at least two players, within a specific time window. Notably, a key indicator of interactivity was the presence of rich, conversational audio. We used WhisperX-generated audio transcripts to create a dataset and identify these interactions. To preserve conversational coherence, we split the transcript into segments such that a new segment begins after a pause of at least 10 seconds between utterances. These segments represent potential areas of interactivity and Minecraft social interactions. Given the scale of the resulting dataset, manual annotation of segments was infeasible. Instead, we leveraged recent advances in large language models to perform automated annotation. A binary classification approach using a 4-shot prompt demonstrated high accuracy and efficiency on a validation subset. Based on this evaluation and overall cost considerations, we used DeepSeek-484
V3.2-fast~\citep{deepseekai2025deepseekv32pushingfrontieropen} to label the full dataset with binary annotations. Table~\ref{tab:labels_questions} demonstrates the labeling scheme. Segments assigned specific labels were then identified as highly interactive and selected for further analysis. 

\newpage
\onecolumn 
\subsection{Additional Dataset Visual Examples}

We demonstrate more visual examples in \cref{fig:appendix-visual-examples-1}, \cref{fig:appendix-visual-examples-2}, \cref{fig:appendix-visual-examples-3} of the dataset. The visualization follows the same format as \cref{fig:dataset-dynamic-plot}, but here we also depict temporal progression: frames within each row correspond to a single example, are spaced 1 second apart, and progress from left to right in time.

\begin{figure}[H]
  \centering
  \includegraphics[width=\textwidth]{assets/figs/appendix_visual_1.pdf}
  \caption{
    Dataset visual examples. 
  }
  \label{fig:appendix-visual-examples-1}
\end{figure}

\begin{figure}[H]
  \centering
  \includegraphics[width=\textwidth]{assets/figs/dante-morgan-xander.pdf}
  \caption{
    Dataset visual examples. 
  }
  \label{fig:appendix-visual-examples-2}
\end{figure}

\begin{figure}[H]
  \centering
  \includegraphics[width=\textwidth]{assets/figs/appendix_visual_2.pdf}
  \caption{
    Dataset visual examples. 
  }
  \label{fig:appendix-visual-examples-3}
\end{figure}




\newpage
\subsection{Complete Evaluation Prompt Table}
\label{appendix_details_of_evaluation}
\scriptsize
\begingroup
\scriptsize
\setlength{\tabcolsep}{2pt}
\renewcommand{\arraystretch}{0.96}
\everymath{\scriptstyle}
\newcolumntype{Y}{>{\raggedright\arraybackslash}X}
\newcolumntype{C}[1]{>{\centering\arraybackslash}p{#1}}
\newcolumntype{L}[1]{>{\raggedright\arraybackslash}p{#1}}

\begin{xltabular}{\textwidth}{@{}
  C{0.035\textwidth}
  L{0.135\textwidth}
  Y
  L{0.095\textwidth}
  C{0.070\textwidth}
  C{0.045\textwidth}
  C{0.045\textwidth}
  C{0.075\textwidth}
  C{0.075\textwidth}
@{}}
  \caption{Prompt--response evaluation pairs. Prompt consists of the agent's entire lived life up to the start of response. The CHC column refers to the categories in \citet{schneider2018cattell}. All three videos are from anonymous player Morgan, video 1 corresponds to recording session d868fc041c231673, video 2 corresponds to b164ada80b3e43a0, video 3 corresponds to d4e57df96d532e93.}
  \label{tab:full-prompt-table} \\
  \toprule
  \textbf{ID} &
  \textbf{Test type} &
  \textbf{Spoken instruction} &
  \textbf{Response} &
  \textbf{Metric} &
  \textbf{CHC} &
  \textbf{Video} &
  \textbf{$R_{\text{start}}$ (ms)} &
  \textbf{$R_{\text{end}}$ (ms)} \\
  \midrule
  \endfirsthead

  \caption[]{Prompt--response evaluation pairs, continued.} \\
  \toprule
  \textbf{ID} &
  \textbf{Test type} &
  \textbf{Spoken instruction} &
  \textbf{Response} &
  \textbf{Metric} &
  \textbf{CHC} &
  \textbf{Video} &
  \textbf{$R_{\text{start}}$ (ms)} &
  \textbf{$R_{\text{end}}$ (ms)} \\
  \midrule
  \endhead

  \midrule
  \multicolumn{9}{r}{\footnotesize Continued on next page} \\
  \endfoot

  \bottomrule
  \endlastfoot

  1  & audio to action binding & Please press Space. & Space pressed. & $\mathrm{D}_{\mathit{K}}$ & $Ga$ & 1 & 67970  & 68120 \\
  2  & audio to action binding & Please press Shift. & Shift pressed. & $\mathrm{D}_{\mathit{K}}$ & $Ga$ & 1 & 73120  & 73650 \\
  3  & audio to action binding & Please press T. & T pressed. & $\mathrm{D}_{\mathit{K}}$ & $Ga$ & 1 & 89240  & 89410 \\
  4  & spelling ability & Please press the letter that ends the word STOP, as in S-T-O... & P pressed. & $\mathrm{D}_{\mathit{K}}$ & $Grw$ & 1 & 115950 & 116070 \\
  5  & spelling ability & Please press the letter that ends big, as in B-I... & G pressed. & $\mathrm{D}_{\mathit{K}}$ & $Grw$ & 1 & 136930 & 137050 \\
  6  & audio to action binding & Please press the letter Q. & Q pressed. & $\mathrm{D}_{\mathit{K}}$ & $Ga$ & 1 & 161850 & 162020 \\
  7  & action recognition & What did you just close? & Door. & $\mathrm{Acc}_{\mathit{R}}$ & $Gv$ & 1 & 200116 & 200550 \\
  8  & internal state recognition & What is that green dot? & XP. & $\mathrm{Acc}_{\mathit{R}}$ & $Gf$ & 1 & 435183 & 435866 \\
  9  & world model understanding & I'm going to shoot you with an arrow, okay? And I want you to tell me how many hearts of damage are done, okay? & Two. & $\mathrm{Acc}_{\mathit{R}}$ & $Gf$ & 1 & 482750 & 483500 \\
  10 & object recognition & What do you have in your right hand? & Carrot. & $\mathrm{Acc}_{\mathit{R}}$ & $Gv$ & 1 & 510550 & 511016 \\
  11 & object recognition & What do you have in your left hand? & Torch. & $\mathrm{Acc}_{\mathit{R}}$ & $Gv$ & 1 & 514683 & 515383 \\
  12 & color understanding & What color sheep am I standing next to? & Black. & $\mathrm{Acc}_{\mathit{R}}$ & $Gv$ & 1 & 534633 & 535216 \\
  13 & color understanding & What color sheep am I standing next to? & Gray. & $\mathrm{Acc}_{\mathit{R}}$ & $Gv$ & 1 & 548100 & 548566 \\
  14 & color understanding & What color flower am I standing next to and looking at? & White. & $\mathrm{Acc}_{\mathit{R}}$ & $Gv$ & 1 & 558866 & 559500 \\
  15 & color understanding & What color of flower am I looking at? & Red. & $\mathrm{Acc}_{\mathit{R}}$ & $Gv$ & 1 & 566966 & 567433 \\
  16 & color understanding & What color of flower am I looking at? & Yellow. & $\mathrm{Acc}_{\mathit{R}}$ & $Gv$ & 1 & 572383 & 572899 \\
  17 & type recognition & What is the stack of stuff that I'm standing next to made out of? & Wool. & $\mathrm{Acc}_{\mathit{R}}$ & $Gv$ & 1 & 1111099 & 1111616 \\
  18 & arithmetic & If you add up the number of blocks in the wool pile and the number of blocks in the granite pile, how many blocks total are there? & Five. & $\mathrm{Acc}_{\mathit{R}}$ & $Gq$ & 1 & 1182683 & 1183299 \\
  19 & arithmetic & Now, if you add up the number of blocks in the granite pile and the wool pile, how many blocks are there total? & Four. & $\mathrm{Acc}_{\mathit{R}}$ & $Gq$ & 1 & 1199099 & 1199599 \\
  20 & arithmetic & I am going to place a block in the granite pile. How many blocks are in the granite pile now? & Three. & $\mathrm{Acc}_{\mathit{R}}$ & $Gq$ & 1 & 1235883 & 1236333 \\
  21 & object constancy & Now, jump up. Can you see me? & Yes. & $\mathrm{Acc}_{\mathit{R}}$ & $Gv$ & 2 & 605866 & 606200 \\
  22 & object constancy & Jump up. Did I move? & No. & $\mathrm{Acc}_{\mathit{R}}$ & $Gv$ & 2 & 615450 & 615899 \\
  23 & audio to action binding & Press W. & W pressed. & $\mathrm{D}_{\mathit{K}}$ & $Ga$ & 2 & 679650 & 679740 \\
  24 & audio to action binding & Press S. & S pressed. & $\mathrm{D}_{\mathit{K}}$ & $Ga$ & 2 & 684010 & 684390 \\
  25 & audio to action binding & Press A. & A pressed. & $\mathrm{D}_{\mathit{K}}$ & $Ga$ & 2 & 689700 & 689850 \\
  26 & audio to action binding & Press D. & D pressed. & $\mathrm{D}_{\mathit{K}}$ & $Ga$ & 2 & 692750 & 692940 \\
  27 & audio to action binding & Press S. & S pressed. & $\mathrm{D}_{\mathit{K}}$ & $Ga$ & 2 & 697070 & 697150 \\
  28 & audio to action binding & Do it again. Press S. & S pressed. & $\mathrm{D}_{\mathit{K}}$ & $Ga$ & 2 & 709500 & 709700 \\
  29 & audio to action binding & Press W. & W pressed. & $\mathrm{D}_{\mathit{K}}$ & $Ga$ & 2 & 713210 & 713440 \\
  30 & working memory, object constancy & Have you been here before? & Yes. & $\mathrm{Acc}_{\mathit{R}}$ & $Gsm$ & 2 & 1456133 & 1456466 \\
  31 & working memory, object constancy & Will we be someplace we've been before or will we be someplace new? So say old or new. & Old. & $\mathrm{Acc}_{\mathit{R}}$ & $Gsm$ & 2 & 1796900 & 1797250 \\
  32 & object recognition & Can you see a spider? & Yes. & $\mathrm{Acc}_{\mathit{R}}$ & $Gv$ & 2 & 1852349 & 1852683 \\
  33 & object recognition & Can you see any sheep? & Yes. & $\mathrm{Acc}_{\mathit{R}}$ & $Gv$ & 2 & 1857016 & 1857549 \\
  34 & type recognition & What kind of block am I standing on? & Sand. & $\mathrm{Acc}_{\mathit{R}}$ & $Gv$ & 2 & 1926616 & 1927033 \\
  35 & object recognition & What am I in right now? & Water. & $\mathrm{Acc}_{\mathit{R}}$ & $Gv$ & 2 & 1940433 & 1940783 \\
  36 & world model understanding & Is the sheep dead? & No. & $\mathrm{Acc}_{\mathit{R}}$ & $Gf$ & 3 & 103183 & 103466 \\
  37 & world model understanding & Is the sheep dead? & Yes. & $\mathrm{Acc}_{\mathit{R}}$ & $Gf$ & 3 & 110816 & 111266 \\
  38 & symbol recognition, numerosity & How many pieces of wool are in the box? & Four. & $\mathrm{Acc}_{\mathit{R}}$ & $Gq$ & 3 & 319983 & 320266 \\
  39 & symbol recognition, numerosity & How many different kinds of things are in the box? & Two. & $\mathrm{Acc}_{\mathit{R}}$ & $Gq$ & 3 & 324983 & 325166 \\
  40 & symbol recognition, numerosity & How many pieces of wood are in the box? & Twelve. & $\mathrm{Acc}_{\mathit{R}}$ & $Gq$ & 3 & 346466 & 346983 \\
  41 & type recognition & What is the first thing that I put in the box? & Wood. & $\mathrm{Acc}_{\mathit{R}}$ & $Gsm$ & 3 & 487899 & 488516 \\
  42 & spatial understanding & Approximately how many blocks away is the box? & Hundred. & $\mathrm{Acc}_{\mathit{R}}$ & $Gv$ & 3 & 583549 & 584283 \\
  43 & object recognition & Morgan, in front of me, what is that? & Zombie. & $\mathrm{Acc}_{\mathit{R}}$ & $Gv$ & 3 & 849333 & 849750 \\
  44 & object recognition & Morgan, are those torches? & Yeah. & $\mathrm{Acc}_{\mathit{R}}$ & $Gv$ & 3 & 895549 & 896200 \\
  45 & symbol recognition, numerosity & How many pieces of wood? & Six. & $\mathrm{Acc}_{\mathit{R}}$ & $Gq$ & 3 & 1669933 & 1670616 \\
  46 & reading & What is the name of the last person who left the game? & Brent. & $\mathrm{Acc}_{\mathit{R}}$ & $Grw$ & 3 & 1908099 & 1908866 \\
  47 & reading & Okay, what is the name of the person who joined just after me? & Chloe. & $\mathrm{Acc}_{\mathit{R}}$ & $Grw$ & 3 & 1942033 & 1942466 \\
  48 & long term memory & What is the middle item in the chest? & Carrot. & $\mathrm{Acc}_{\mathit{R}}$ & $Glr$ & 3 & 2174783 & 2175283 \\
\end{xltabular}
\endgroup
\normalsize

\newpage
\subsection{Evaluation Prompt Visual Examples}

We pick a few prompts and show the nearest frames before response for visualization in Table~\ref{tab:visualization_prompt}.

\begingroup
\footnotesize
\renewcommand{\arraystretch}{1.15}
\setlength{\tabcolsep}{6pt}

\begin{xltabular}{\textwidth}{
  >{\centering\arraybackslash}m{0.55\textwidth}
  >{\raggedright\arraybackslash}X
  >{\centering\arraybackslash}m{1.9cm}
}
  \caption{Prompt visualization examples.}
  \label{tab:visualization_prompt} \\
  \toprule
  \textbf{Nearest Frame} & \textbf{Spoken Instruction} & \textbf{Response} \\
  \midrule
  \endfirsthead

  \caption[]{Prompt visualization examples (continued).} \\
  \toprule
  \textbf{Nearest Frame} & \textbf{Spoken Instruction} & \textbf{Response} \\
  \midrule
  \endhead

  \midrule
  \multicolumn{3}{r}{\footnotesize Continued on next page} \\
  \endfoot

  \bottomrule
  \endlastfoot

  \includegraphics[width=\linewidth,height=3.3cm,keepaspectratio]{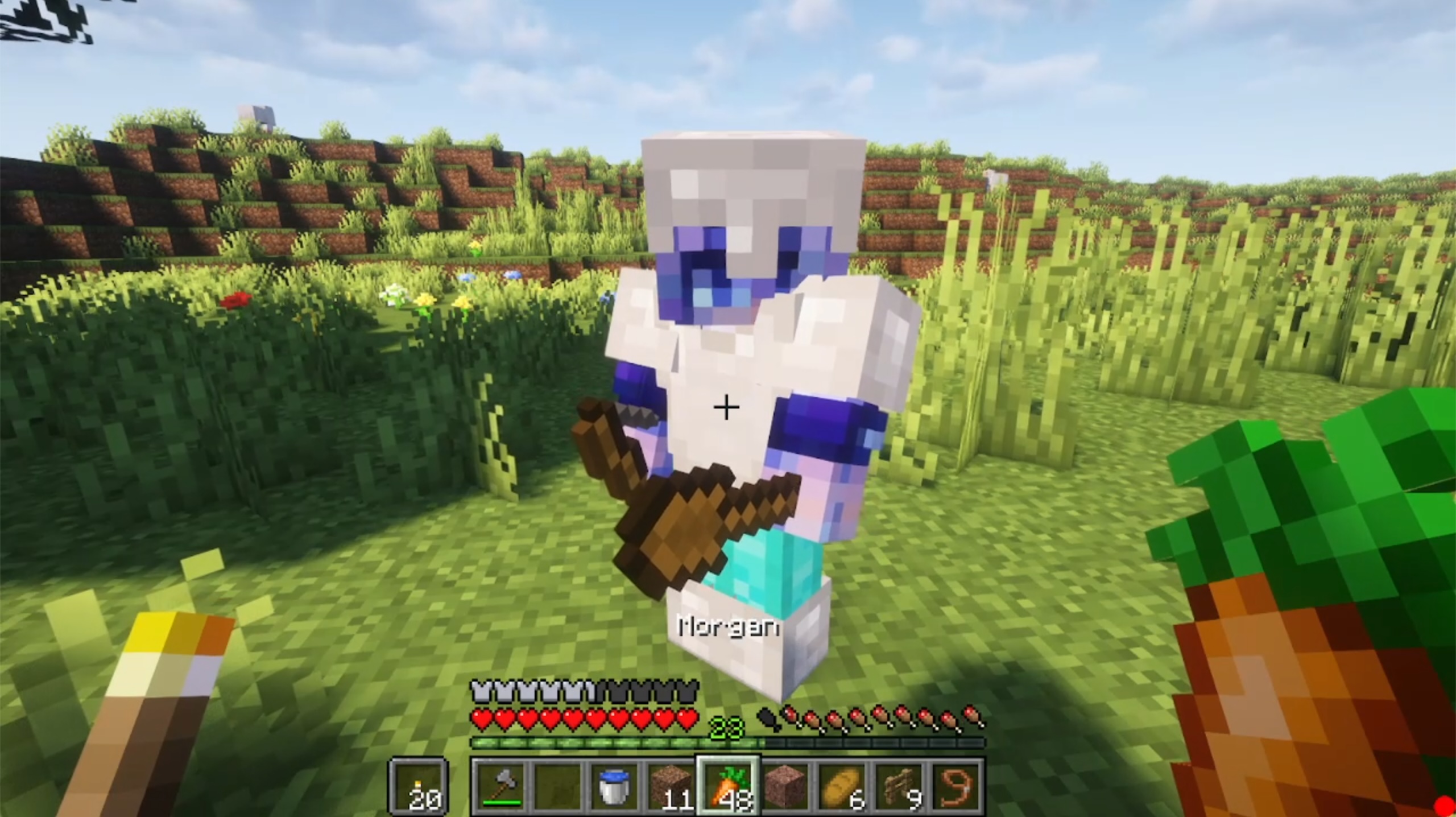} &
  What do you have in your right hand? &
  Carrot. \\
  \midrule

  \includegraphics[width=\linewidth,height=3.3cm,keepaspectratio]{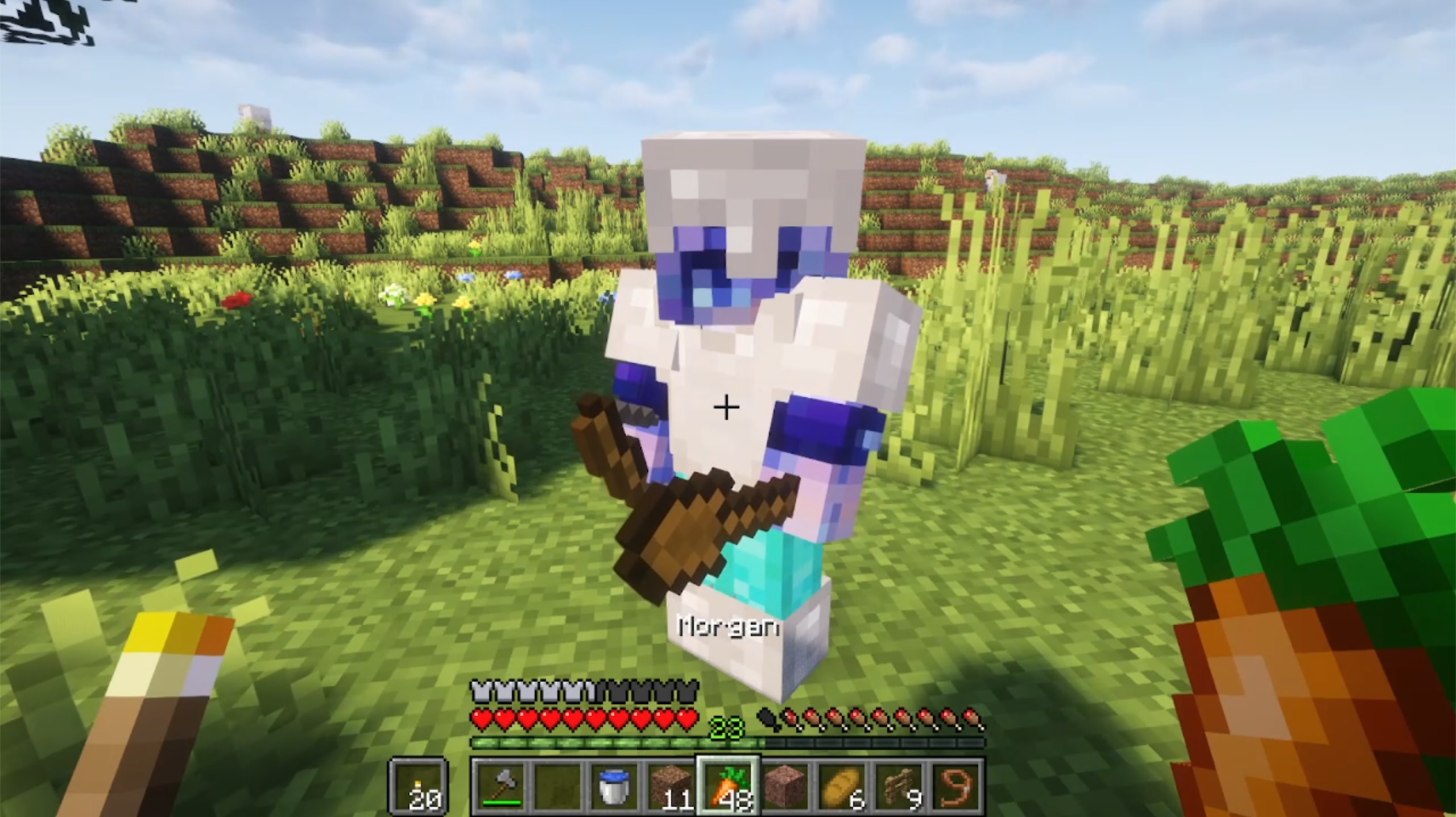} &
  What do you have in your left hand? &
  Torch. \\
  \midrule

  \includegraphics[width=\linewidth,height=3.3cm,keepaspectratio]{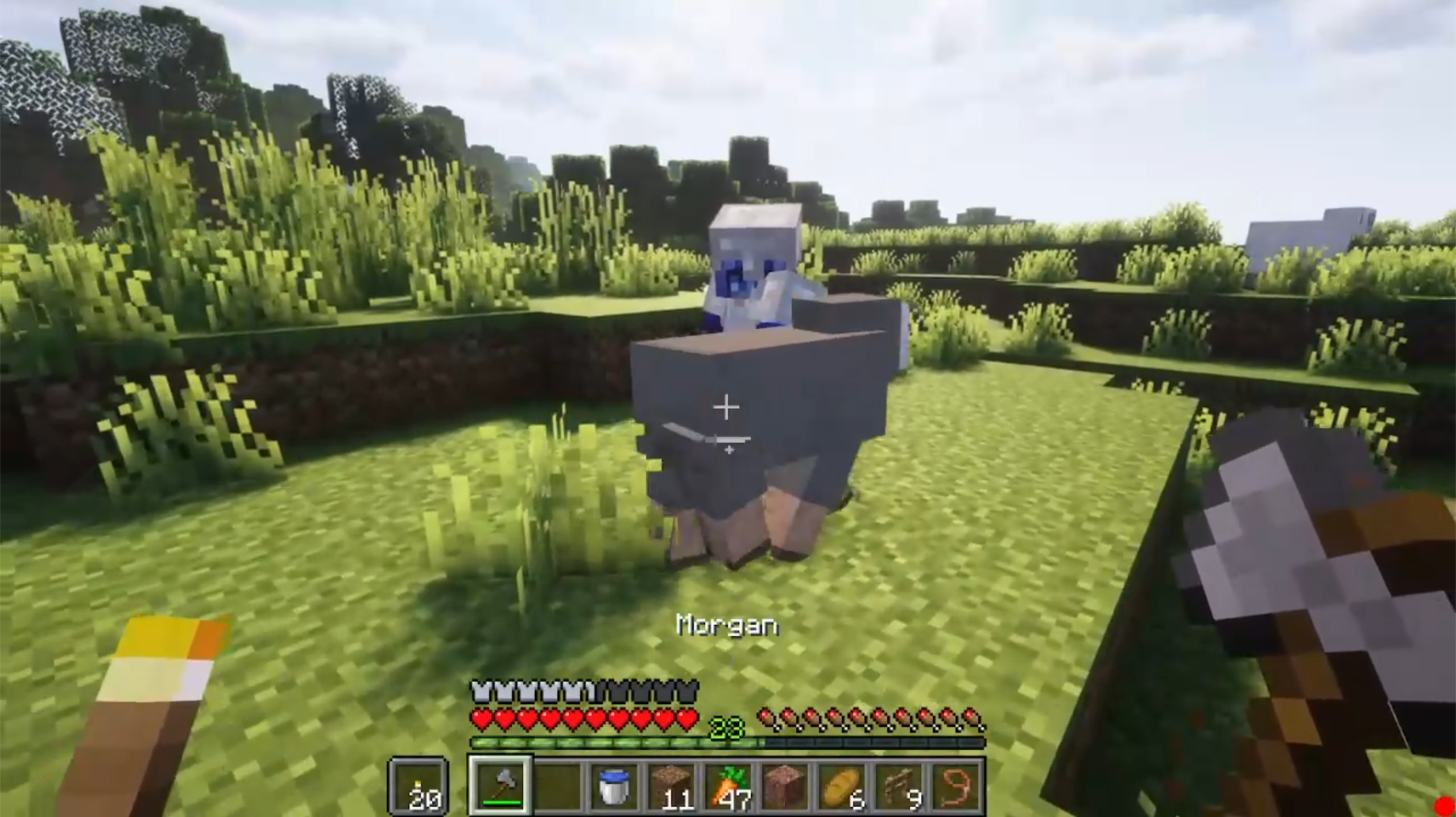} &
  What color sheep am I standing next to? &
  Gray. \\
  \midrule

  \includegraphics[width=\linewidth,height=3.3cm,keepaspectratio]{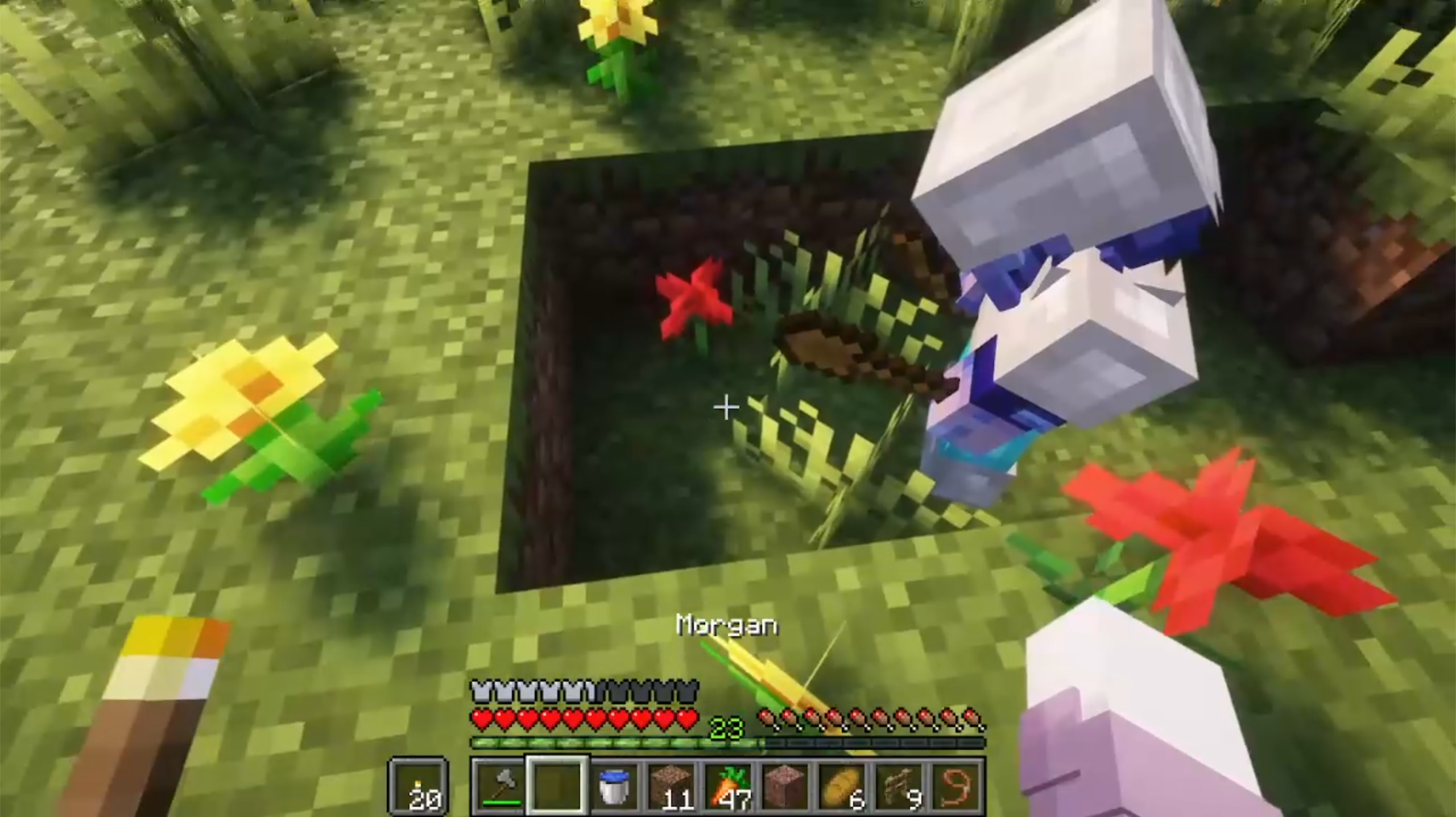} &
  What color of flower am I looking at? &
  Red. \\
  \midrule

  \includegraphics[width=\linewidth,height=3.3cm,keepaspectratio]{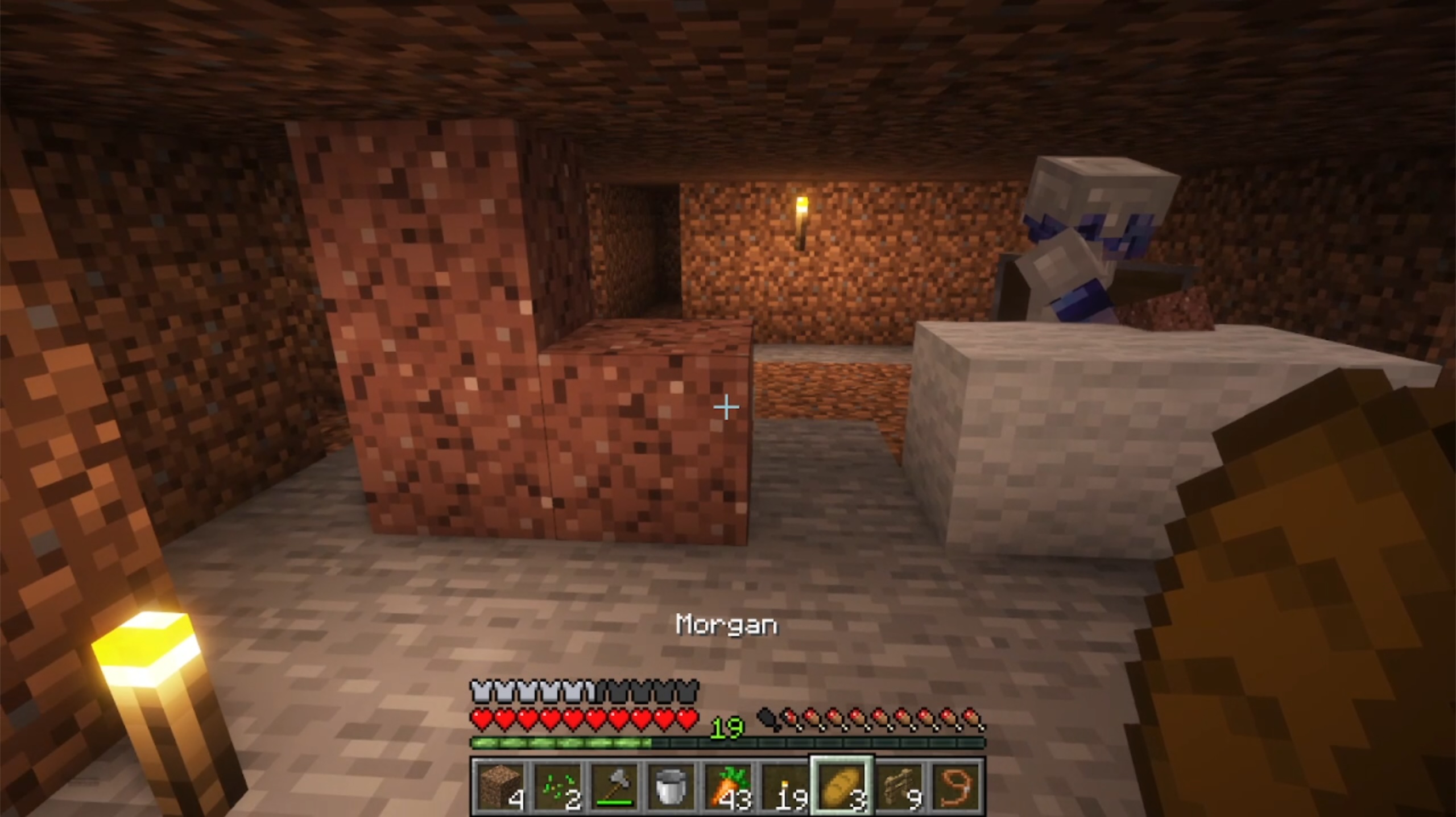} &
  What is the stack of stuff that I'm standing next to made out of? &
  Wool. \\
  \midrule

  \includegraphics[width=\linewidth,height=3.3cm,keepaspectratio]{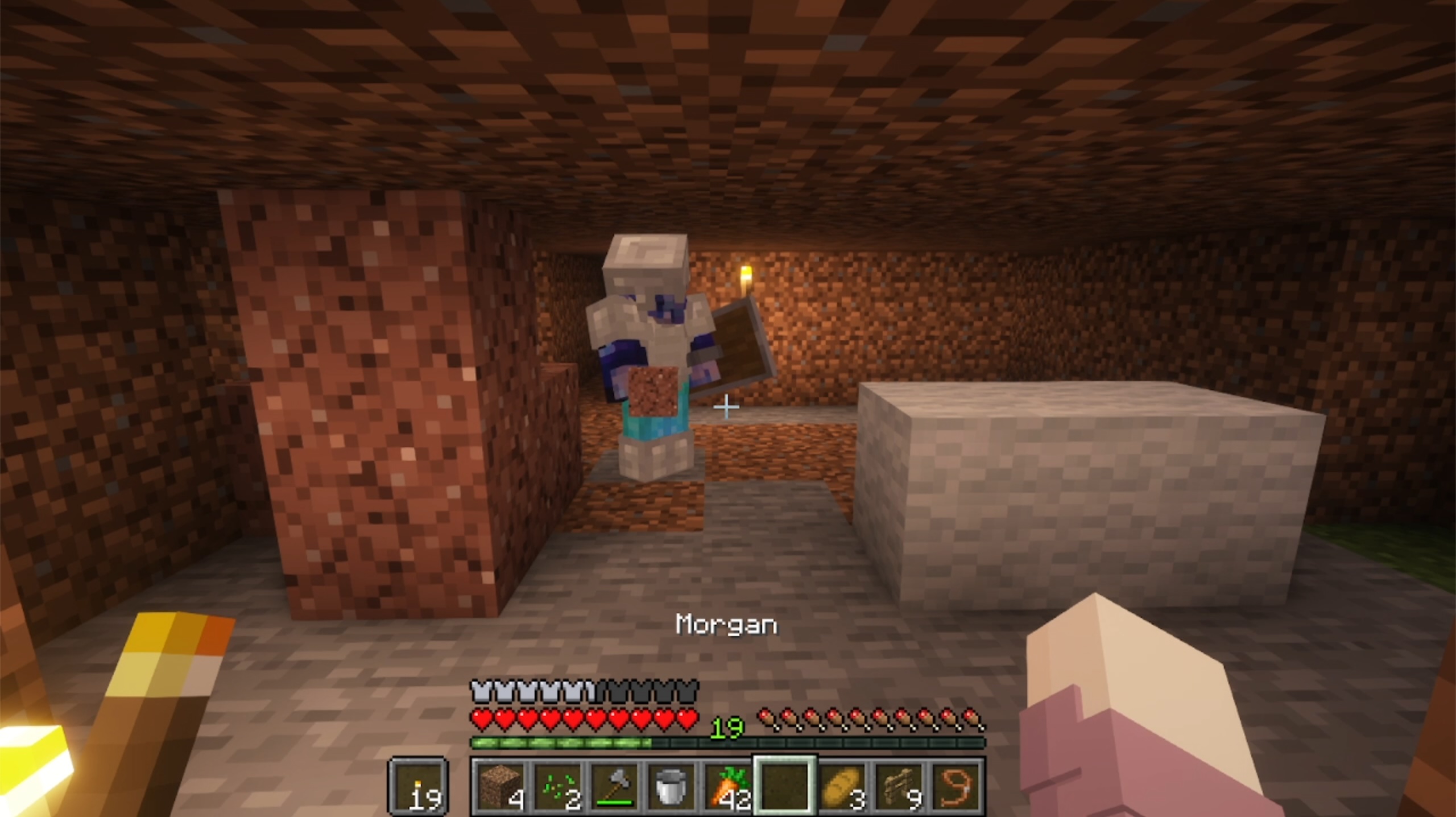} &
  I am going to place a block in the granite pile. How many blocks are in the granite pile now? &
  Three. \\
  \midrule

  \includegraphics[width=\linewidth,height=3.3cm,keepaspectratio]{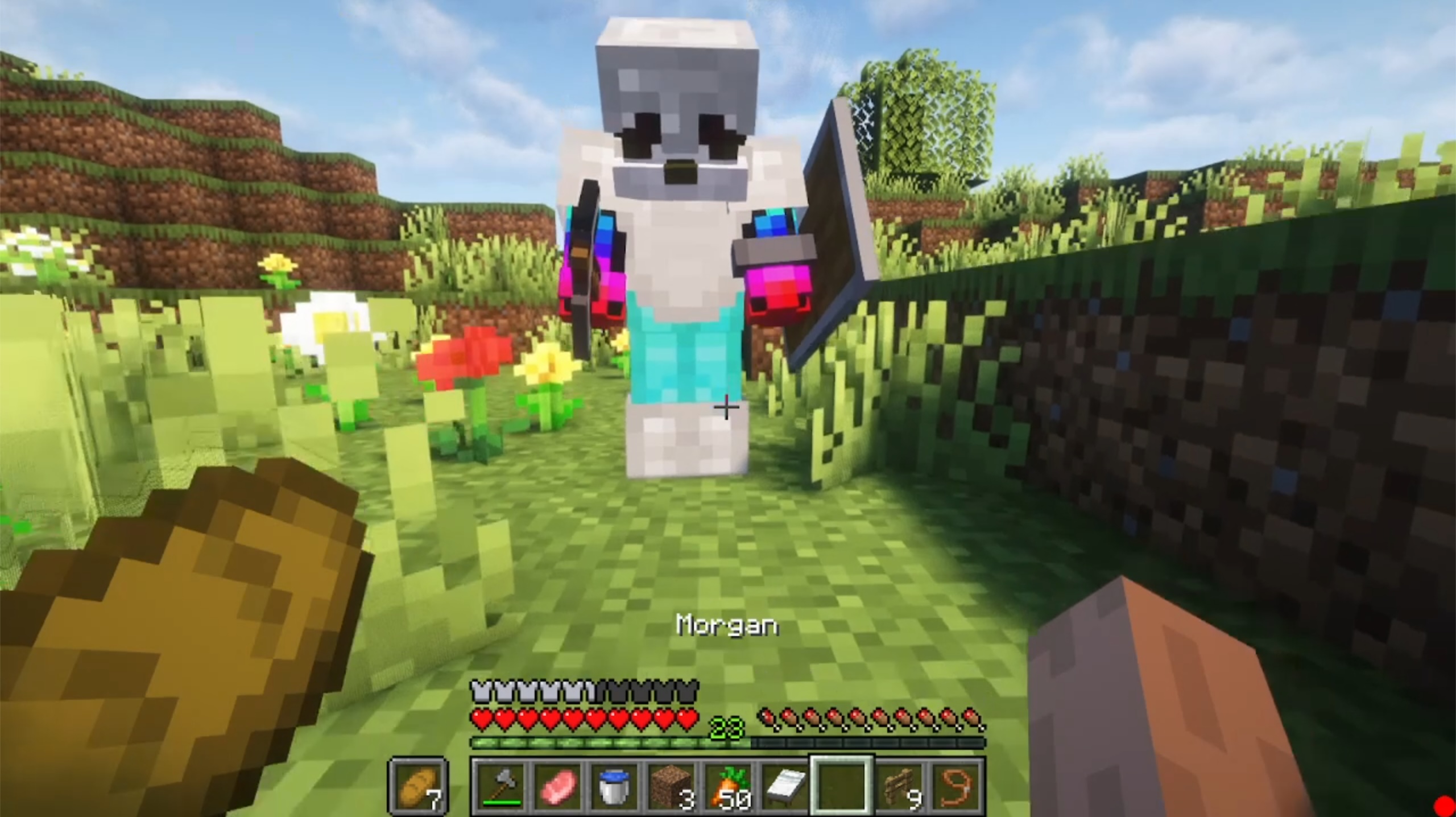} &
  Now, jump up. Can you see me? &
  Yes. \\
  \midrule

  \includegraphics[width=\linewidth,height=3.3cm,keepaspectratio]{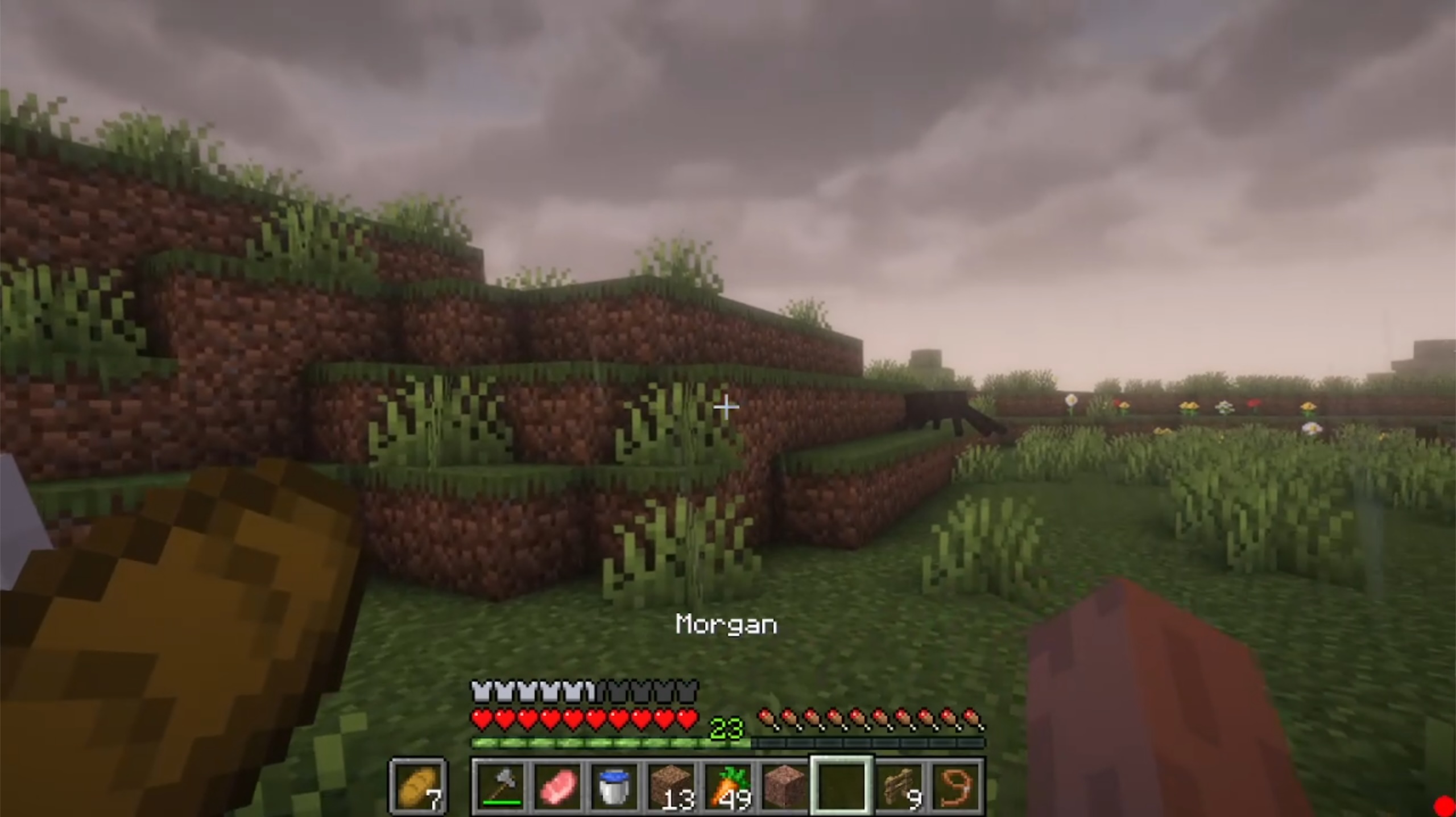} &
  Can you see a spider? &
  Yes. \\
  \midrule

  \includegraphics[width=\linewidth,height=3.3cm,keepaspectratio]{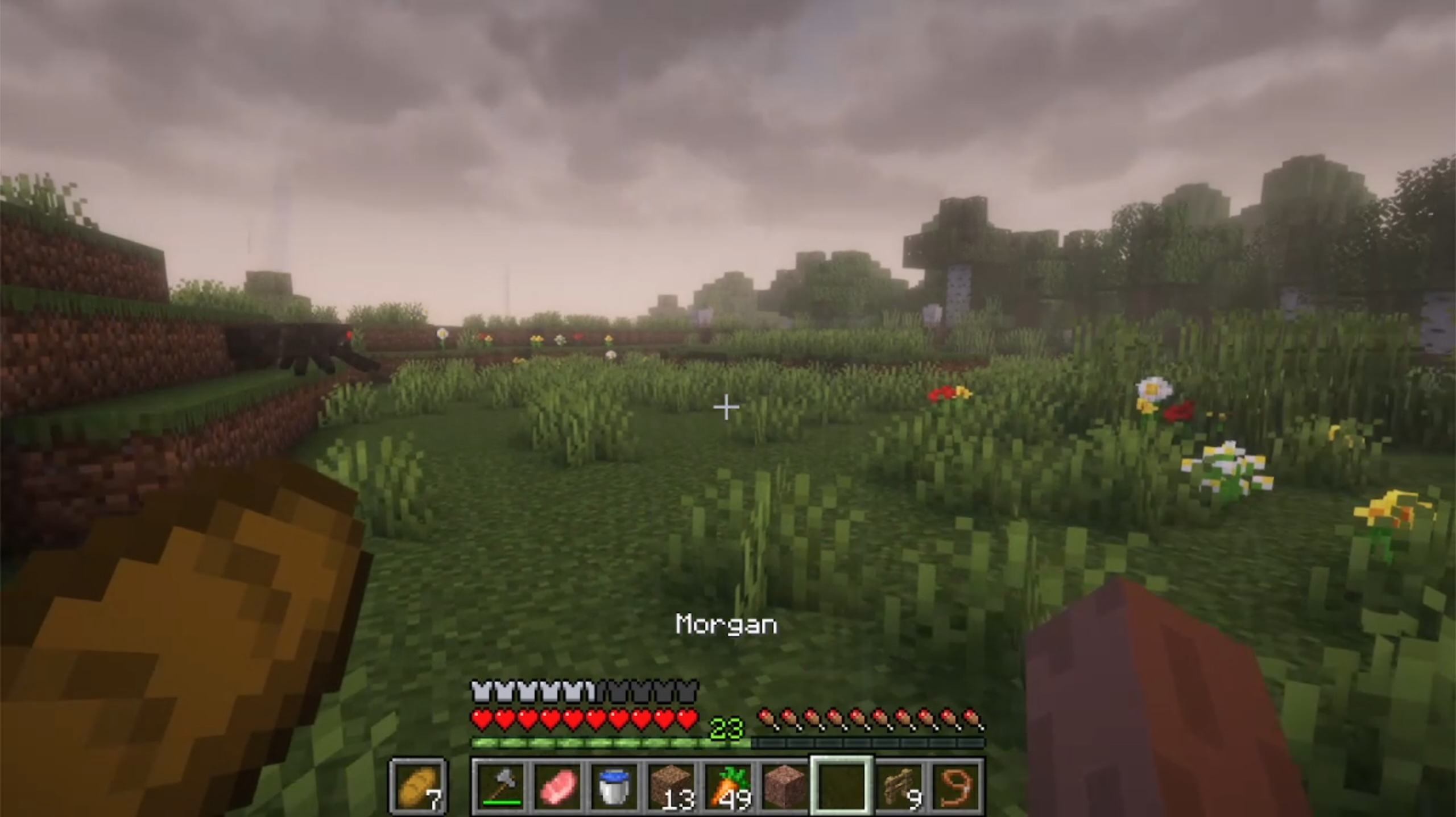} &
  Can you see any sheep? &
  Yes. \\
  \midrule

  \includegraphics[width=\linewidth,height=3.3cm,keepaspectratio]{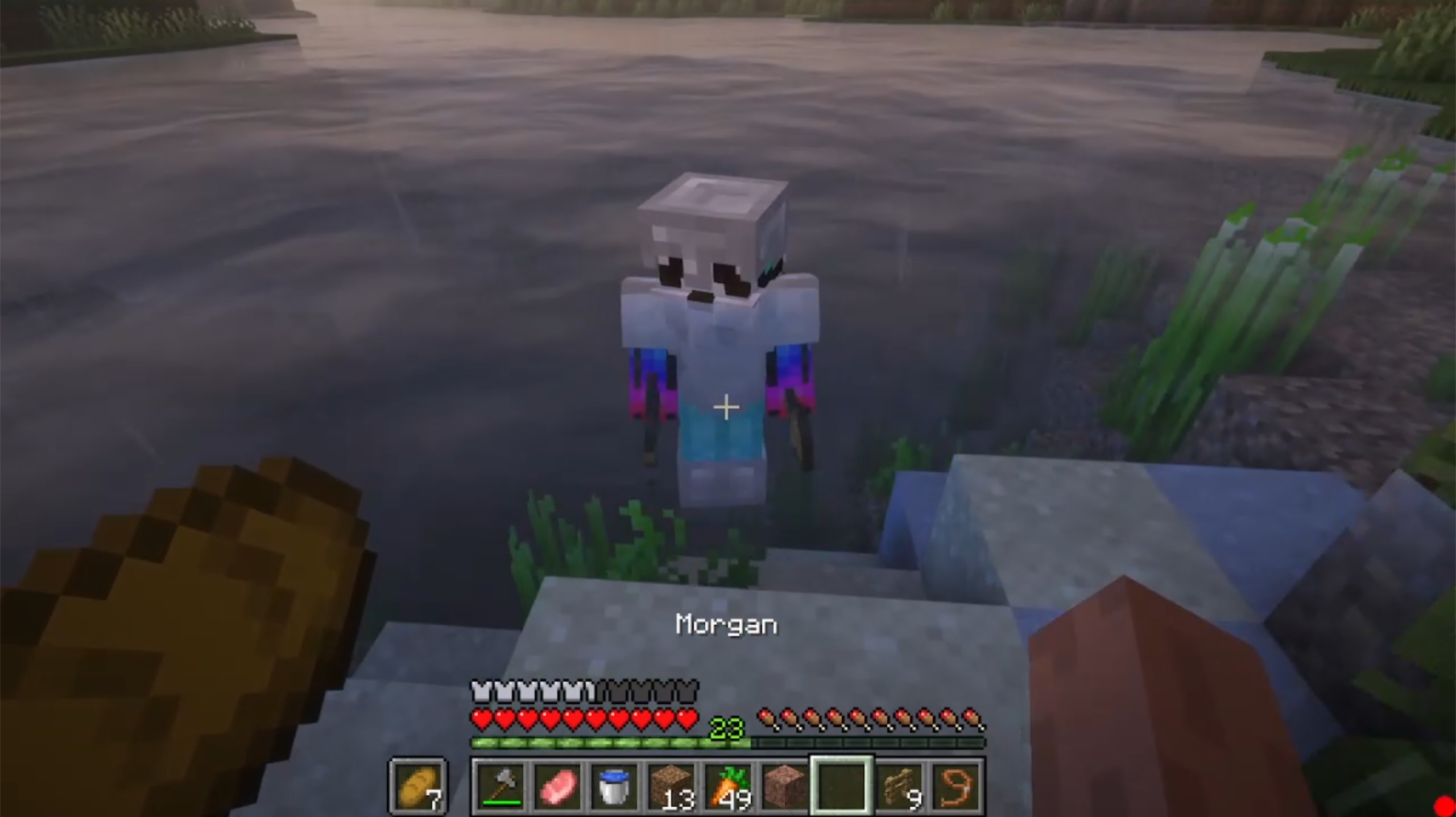} &
  What am I in right now? &
  Water. \\

\end{xltabular}
\endgroup

\end{document}